\PassOptionsToPackage{hyphens}{url}
\documentclass[withtimes]{easychair}

\usepackage[T1]{fontenc}
\usepackage[utf8]{inputenc}
\usepackage{cmtt}

\usepackage{amsmath,amssymb,amsfonts,mathtools}
\usepackage{amsthm}
\usepackage{array}
\usepackage{booktabs}
\usepackage{multirow}
\usepackage{siunitx}
\usepackage{xcolor}
\usepackage{graphicx}
\usepackage{wrapfig}
\usepackage{subcaption}
\usepackage{float}
\usepackage{placeins}
\usepackage{enumitem}
\usepackage{algorithm}
\usepackage[noend]{algpseudocode}
\usepackage{listings}
\usepackage{upquote}
\usepackage{pmboxdraw}
\usepackage{pifont}
\usepackage{nicefrac}
\usepackage{microtype}
\usepackage[skins,breakable]{tcolorbox}
\usepackage[numbers,sort&compress]{natbib}

\AtBeginEnvironment{algorithmic}{\small}
\algrenewcommand\algorithmicindent{1em}
\tcbset{boxsep=1pt, top=2pt, bottom=2pt, left=3pt, right=3pt, before skip=5pt, after skip=5pt}

\usepackage{tikz}
\usetikzlibrary{
  positioning, fit, calc,
  arrows.meta,
  decorations.pathmorphing
}

\definecolor{keywordcolor}{rgb}{0.7, 0.1, 0.1}
\definecolor{tacticcolor}{rgb}{0.0, 0.1, 0.6}
\definecolor{commentcolor}{rgb}{0.4, 0.4, 0.4}
\definecolor{symbolcolor}{rgb}{0.0, 0.1, 0.6}
\definecolor{codebg}{gray}{0.60}
\definecolor{sortcolor}{rgb}{0.0, 0.4, 0.0}
\definecolor{attributecolor}{rgb}{0.5, 0.2, 0.5}

\tikzset{
  hand/.style={
    draw,
    rounded corners=6pt,
    line width=0.9pt,
    decorate,
    decoration={random steps,segment length=8pt,amplitude=0.6pt},
    fill=white
  },
  handthin/.style={
    draw,
    rounded corners=6pt,
    line width=0.7pt,
    decorate,
    decoration={random steps,segment length=7pt,amplitude=0.5pt},
    fill=white
  },
  hlabel/.style={font=\small},
  box/.style={hand, align=left, inner sep=6pt, font=\small},
  bigbox/.style={hand, align=center, inner sep=10pt, font=\small},
  arrow/.style={-{Stealth[length=3mm,width=2mm]}, line width=0.9pt},
  note/.style={font=\small}
}

\lstdefinelanguage{lean} {
mathescape=false,
texcl=false,
morekeywords=[1]{
import, prelude, protected, private, noncomputable, definition, meta, renaming,
hiding, parameter, parameters, begin, constant, constants,
lemma, variable, variables, theory,
print, theorem, example,
open, as, export, override, axiom, axioms, inductive, with,
structure, record, universe, universes,
alias, help, precedence, reserve, declare_trace, add_key_equivalence,
match, infix, infixl, infixr, notation, postfix, prefix, instance,
eval, reduce, check, end, this,
using, using_well_founded, namespace, section,
attribute, local, set_option, extends, include, omit, class,
raw, replacing,
calc, have, show, suffices, by, in, at, let, forall, Pi, fun,
exists, if, dif, then, else, assume, obtain, from, register_simp_ext, unless, break, continue,
mutual, do, def, run_cmd, const,
partial, mut, where, macro, syntax, deriving,
return, try, catch, for, macro_rules, declare_syntax_cat, abbrev},
morekeywords=[2]{Sort, Type, Prop},
morekeywords=[3]{
assumption,
apply, intro, intros, allGoals,
generalize, clear, revert, done, exact,
refine, repeat, cases, rewrite, rw,
simp, simp_all, contradiction,
constructor, injection,
induction,
},
literate=
{α}{{\ensuremath{\mathrm{\alpha}}}}1
{β}{{\ensuremath{\mathrm{\beta}}}}1
{γ}{{\ensuremath{\mathrm{\gamma}}}}1
{δ}{{\ensuremath{\mathrm{\delta}}}}1
{ε}{{\ensuremath{\mathrm{\varepsilon}}}}1
{ζ}{{\ensuremath{\mathrm{\zeta}}}}1
{η}{{\ensuremath{\mathrm{\eta}}}}1
{θ}{{\ensuremath{\mathrm{\theta}}}}1
{ι}{{\ensuremath{\mathrm{\iota}}}}1
{κ}{{\ensuremath{\mathrm{\kappa}}}}1
{μ}{{\ensuremath{\mathrm{\mu}}}}1
{ν}{{\ensuremath{\mathrm{\nu}}}}1
{ξ}{{\ensuremath{\mathrm{\xi}}}}1
{π}{{\ensuremath{\mathrm{\mathnormal{\pi}}}}}1
{ρ}{{\ensuremath{\mathrm{\rho}}}}1
{σ}{{\ensuremath{\mathrm{\sigma}}}}1
{τ}{{\ensuremath{\mathrm{\tau}}}}1
{φ}{{\ensuremath{\mathrm{\varphi}}}}1
{χ}{{\ensuremath{\mathrm{\chi}}}}1
{ψ}{{\ensuremath{\mathrm{\psi}}}}1
{ω}{{\ensuremath{\mathrm{\omega}}}}1
{Γ}{{\ensuremath{\mathrm{\Gamma}}}}1
{Δ}{{\ensuremath{\mathrm{\Delta}}}}1
{Θ}{{\ensuremath{\mathrm{\Theta}}}}1
{Λ}{{\ensuremath{\mathrm{\Lambda}}}}1
{Σ}{{\ensuremath{\mathrm{\Sigma}}}}1
{Φ}{{\ensuremath{\mathrm{\Phi}}}}1
{Ξ}{{\ensuremath{\mathrm{\Xi}}}}1
{Ψ}{{\ensuremath{\mathrm{\Psi}}}}1
{Ω}{{\ensuremath{\mathrm{\Omega}}}}1
{ℵ}{{\ensuremath{\aleph}}}1
{≤}{{\ensuremath{\leq}}}1
{≥}{{\ensuremath{\geq}}}1
{≠}{{\ensuremath{\neq}}}1
{≈}{{\ensuremath{\approx}}}1
{≡}{{\ensuremath{\equiv}}}1
{≃}{{\ensuremath{\simeq}}}1
{≤}{{\ensuremath{\leq}}}1
{≥}{{\ensuremath{\geq}}}1
{∂}{{\ensuremath{\partial}}}1
{∆}{{\ensuremath{\triangle}}}1 
{∫}{{\ensuremath{\int}}}1
{∑}{{\ensuremath{\mathrm{\Sigma}}}}1
{Π}{{\ensuremath{\mathrm{\Pi}}}}1
{⊥}{{\ensuremath{\perp}}}1
{∞}{{\ensuremath{\infty}}}1
{∂}{{\ensuremath{\partial}}}1
{∓}{{\ensuremath{\mp}}}1
{±}{{\ensuremath{\pm}}}1
{×}{{\ensuremath{\times}}}1
{⊕}{{\ensuremath{\oplus}}}1
{⊗}{{\ensuremath{\otimes}}}1
{⊞}{{\ensuremath{\boxplus}}}1
{∇}{{\ensuremath{\nabla}}}1
{√}{{\ensuremath{\sqrt}}}1
{⬝}{{\ensuremath{\cdot}}}1
{•}{{\ensuremath{\cdot}}}1
{∘}{{\ensuremath{\circ}}}1
{⁻}{{\ensuremath{^{-}}}}1
{▸}{{\ensuremath{\blacktriangleright}}}1
{∧}{{\ensuremath{\wedge}}}1
{∨}{{\ensuremath{\vee}}}1
{¬}{{\ensuremath{\neg}}}1
{⊢}{{\ensuremath{\vdash}}}1
{⟨}{{\ensuremath{\langle}}}1
{⟩}{{\ensuremath{\rangle}}}1
{↦}{{\ensuremath{\mapsto}}}1
{←}{{\ensuremath{\leftarrow}}}1
{<-}{{\ensuremath{\leftarrow}}}1
{→}{{\ensuremath{\rightarrow}}}1
{↔}{{\ensuremath{\leftrightarrow}}}1
{⇒}{{\ensuremath{\Rightarrow}}}1
{⟹}{{\ensuremath{\Longrightarrow}}}1
{⇐}{{\ensuremath{\Leftarrow}}}1
{⟸}{{\ensuremath{\Longleftarrow}}}1
{∩}{{\ensuremath{\cap}}}1
{∪}{{\ensuremath{\cup}}}1
{⊂}{{\ensuremath{\subseteq}}}1
{⊆}{{\ensuremath{\subseteq}}}1
{⊄}{{\ensuremath{\nsubseteq}}}1
{⊈}{{\ensuremath{\nsubseteq}}}1
{⊃}{{\ensuremath{\supseteq}}}1
{⊇}{{\ensuremath{\supseteq}}}1
{⊅}{{\ensuremath{\nsupseteq}}}1
{⊉}{{\ensuremath{\nsupseteq}}}1
{∈}{{\ensuremath{\in}}}1
{∉}{{\ensuremath{\notin}}}1
{∋}{{\ensuremath{\ni}}}1
{∌}{{\ensuremath{\notni}}}1
{∅}{{\ensuremath{\emptyset}}}1
{∖}{{\ensuremath{\setminus}}}1
{†}{{\ensuremath{\dag}}}1
{ℕ}{{\ensuremath{\mathbb{N}}}}1
{ℤ}{{\ensuremath{\mathbb{Z}}}}1
{ℝ}{{\ensuremath{\mathbb{R}}}}1
{ℚ}{{\ensuremath{\mathbb{Q}}}}1
{ℂ}{{\ensuremath{\mathbb{C}}}}1
{⌞}{{\ensuremath{\llcorner}}}1
{⌟}{{\ensuremath{\lrcorner}}}1
{⦃}{{\ensuremath{\{\!|}}}1
{⦄}{{\ensuremath{|\!\}}}}1
{‖}{{\ensuremath{\|}}}1
{₁}{{\ensuremath{_1}}}1
{₂}{{\ensuremath{_2}}}1
{₃}{{\ensuremath{_3}}}1
{₄}{{\ensuremath{_4}}}1
{₅}{{\ensuremath{_5}}}1
{₆}{{\ensuremath{_6}}}1
{₇}{{\ensuremath{_7}}}1
{₈}{{\ensuremath{_8}}}1
{₉}{{\ensuremath{_9}}}1
{₀}{{\ensuremath{_0}}}1
{ᵢ}{{\ensuremath{_i}}}1
{ⱼ}{{\ensuremath{_j}}}1
{ₐ}{{\ensuremath{_a}}}1
{¹}{{\ensuremath{^1}}}1
{ₙ}{{\ensuremath{_n}}}1
{ₘ}{{\ensuremath{_m}}}1
{ₚ}{{\ensuremath{_p}}}1
{↑}{{\ensuremath{\uparrow}}}1
{↓}{{\ensuremath{\downarrow}}}1
{...}{{\ensuremath{\ldots}}}1
{·}{{\ensuremath{\cdot}}}1
{▸}{{\ensuremath{\triangleright}}}1
{Σ}{{\color{symbolcolor}\ensuremath{\Sigma}}}1
{Π}{{\color{symbolcolor}\ensuremath{\Pi}}}1
{∀}{{\color{symbolcolor}\ensuremath{\forall}}}1
{∃}{{\color{symbolcolor}\ensuremath{\exists}}}1
{λ}{{\color{symbolcolor}\ensuremath{\mathrm{\lambda}}}}1
{\$}{{\color{symbolcolor}\$}}1
{:=}{{\color{symbolcolor}:=}}1
{=}{{\color{symbolcolor}=}}1
{<|>}{{\color{symbolcolor}<|>}}1
{<\$>}{{\color{symbolcolor}<\$>}}1
{+}{{\color{symbolcolor}+}}1
{*}{{\color{symbolcolor}*}}1,
morecomment=[s][\color{commentcolor}]{/-}{-/},
morecomment=[l][\itshape \color{commentcolor}]{--},
showstringspaces=false,
keepspaces=true,
morestring=[b]",
tabsize=3,
extendedchars=false,
sensitive=true,
breaklines=true,
breakatwhitespace=true,
basicstyle=\ttfamily\small,
captionpos=b,
columns=[l]fullflexible,
identifierstyle={\ttfamily\color{black}},
keywordstyle=[1]{\ttfamily\color{keywordcolor}},
keywordstyle=[2]{\ttfamily\color{sortcolor}},
keywordstyle=[3]{\ttfamily\color{tacticcolor}},
keywordstyle=[4]{\ttfamily\color{attributecolor}},
stringstyle=\ttfamily,
commentstyle={\ttfamily\footnotesize },
}

\lstdefinestyle{colored}{
  basicstyle=\ttfamily\small,
  keywordstyle=\color{keywordcolor}\bfseries,
  commentstyle=\color{commentcolor}\itshape,
  stringstyle=\color{symbolcolor},
  columns=fullflexible,
  keepspaces=true,
  showstringspaces=false,
  breaklines=true,
  breakautoindent=false,
  breakindent=0pt,
  upquote=true,
  frame=none,
  backgroundcolor=\color{white},
  xleftmargin=0pt,
  belowskip=0pt,
  moredelim=**[is][\color{tacticcolor}]{@}{@},
}
\lstdefinestyle{simple}{
  basicstyle=\ttfamily\small,
  keywordstyle=\color{black},
  commentstyle=\color{black},
  stringstyle=\color{black},
  columns=fullflexible,
  keepspaces=true,
  showstringspaces=false,
  breaklines=true,
  breakautoindent=false,
  breakindent=0pt,
  upquote=true,
  frame=none,
  backgroundcolor=\color{white},
  xleftmargin=0pt,
  belowskip=0pt,
  moredelim=**[is][\color{black}]{@}{@},
}

\DeclareUnicodeCharacter{00A0}{~}
\DeclareUnicodeCharacter{2018}{`}
\DeclareUnicodeCharacter{2019}{'}
\DeclareUnicodeCharacter{201C}{``}
\DeclareUnicodeCharacter{201D}{''}
\DeclareUnicodeCharacter{2013}{--}
\DeclareUnicodeCharacter{2014}{---}
\DeclareUnicodeCharacter{00D7}{\ensuremath{\times}}
\DeclareUnicodeCharacter{0107}{\'c}
\DeclareUnicodeCharacter{0159}{\v r}
\DeclareUnicodeCharacter{03B1}{$\alpha$}
\DeclareUnicodeCharacter{03B2}{$\beta$}
\DeclareUnicodeCharacter{03B3}{$\gamma$}
\DeclareUnicodeCharacter{03C0}{$\pi$}
\DeclareUnicodeCharacter{03BB}{$\lambda$}
\DeclareUnicodeCharacter{03A3}{$\Sigma$}
\DeclareUnicodeCharacter{03A0}{$\Pi$}
\DeclareUnicodeCharacter{2200}{$\forall$}
\DeclareUnicodeCharacter{2203}{$\exists$}
\DeclareUnicodeCharacter{2208}{$\in$}
\DeclareUnicodeCharacter{2209}{$\notin$}
\DeclareUnicodeCharacter{2227}{$\wedge$}
\DeclareUnicodeCharacter{2228}{$\vee$}
\DeclareUnicodeCharacter{2260}{$\ne$}
\DeclareUnicodeCharacter{2264}{$\le$}
\DeclareUnicodeCharacter{2265}{$\ge$}
\DeclareUnicodeCharacter{2190}{$\leftarrow$}
\DeclareUnicodeCharacter{2192}{$\to$}
\DeclareUnicodeCharacter{2194}{$\leftrightarrow$}
\DeclareUnicodeCharacter{21A6}{$\mapsto$}
\DeclareUnicodeCharacter{2115}{$\mathbb{N}$}
\DeclareUnicodeCharacter{2124}{$\mathbb{Z}$}
\DeclareUnicodeCharacter{211D}{$\mathbb{R}$}
\DeclareUnicodeCharacter{2080}{\ensuremath{{}_0}}
\DeclareUnicodeCharacter{2081}{\ensuremath{{}_1}}
\DeclareUnicodeCharacter{2082}{\ensuremath{{}_2}}
\DeclareUnicodeCharacter{2083}{\ensuremath{{}_3}}
\DeclareUnicodeCharacter{2084}{\ensuremath{{}_4}}
\DeclareUnicodeCharacter{2085}{\ensuremath{{}_5}}
\DeclareUnicodeCharacter{2086}{\ensuremath{{}_6}}
\DeclareUnicodeCharacter{2087}{\ensuremath{{}_7}}
\DeclareUnicodeCharacter{2088}{\ensuremath{{}_8}}
\DeclareUnicodeCharacter{2089}{\ensuremath{{}_9}}
\DeclareUnicodeCharacter{2191}{\ensuremath{\uparrow}}
\DeclareUnicodeCharacter{2211}{\ensuremath{\sum}}
\DeclareUnicodeCharacter{22A2}{\ensuremath{\vdash}}
\DeclareUnicodeCharacter{27E8}{\ensuremath{\langle}}
\DeclareUnicodeCharacter{27E9}{\ensuremath{\rangle}}
\DeclareUnicodeCharacter{1D4DD}{\ensuremath{\mathcal{N}}}

\theoremstyle{definition}
\newtheorem{example}{Example}
\theoremstyle{plain}

\def\framework{$\mathsf{ECP}$}
\newcounter{excnt}

\title{Formally Solving Answer-Construction Problems in Lean}

\author{
Jialiang Sun\inst{1}
\and
Yuzhi Tang\inst{1}
\and
Ao Li\inst{1}
\and
Chris Maddison\inst{1}
\and
Kuldeep Meel\inst{1,2}
}

\institute{
University of Toronto, Toronto, Canada\\
\email{jialiang.sun@mail.utoronto.ca, meel@cs.toronto.edu}
\and
Georgia Institute of Technology, Atlanta, USA
}

\authorrunning{J. Sun et al.}
\titlerunning{Formally Solving Answer-Construction Problems in Lean}

\begin{document}

\maketitle

\begin{abstract}
Large language models (LLMs) have achieved remarkable progress in formal mathematical reasoning.
Mathematical competition problems fall into two broad types: theorem-proving problems ask for a proof of a fully specified statement, whereas answer-construction problems ask the solver to construct an answer object and prove that it satisfies the stated specification. Existing mathematical reasoning engines mainly target theorem-proving problems, yet answer-construction problems remain less studied.
This setting is challenging because model capabilities are misaligned, with general LLMs better suited to answer construction and prover LLMs better suited to proof generation, and because Lean proof checking alone does not rule out inadmissible circular witnesses. To close this gap, we introduce \textit{Enumerate-Conjecture-Prove} (\framework{}), a neuro-symbolic framework for solving answer-construction problems in Lean. \framework{} uses general LLMs to perform bounded enumeration and construct candidate answers, and invokes prover LLMs to produce machine-checked proofs. \framework{} introduces admissibility checking to ensure that each answer is canonical and does not involve a circular argument. On answer-construction problems from PutnamBench and autoformalized MathArena, \framework{} formally solves 17/346 PutnamBench instances and 18/75 MathArena instances with admissible answers and proofs, outperforming LLM baselines at aligned inference budgets. Our code is available at \href{https://github.com/sunjia72/ecp-lpar}{https://github.com/sunjia72/ecp-lpar}.
\end{abstract}

\section{Introduction}
\label{sec:intro}
Large language models (LLMs) have achieved remarkable progress in mathematical reasoning, with broad applications in math education, verified code generation~\citep{dl4tp,yang2024formal}, and research-level math discovery~\citep{novikov2025alphaevolve,tsoukalas2026advancing,sothanaphan2026resolution,surina2026improved}. On one hand, \textit{general LLMs}, such as models in the GPT-5 family, provide powerful human-like \textit{informal math} capabilities. On the other hand, there is a growing community effort around large-scale
formalized mathematics frameworks such as Lean Mathlib~\citep{lean,mathlib}, which provide the ecosystem and training corpus for \textit{prover LLMs} that generate proofs in a machine-verifiable format. State-of-the-art prover LLMs can match and even outperform top human experts in formally solving high-school and undergraduate math competition problems~\citep{alphaproof,chen2025seed,aristotle}.

\begin{figure}[ht!]
    \centering
    \includegraphics[width=1.0\linewidth]{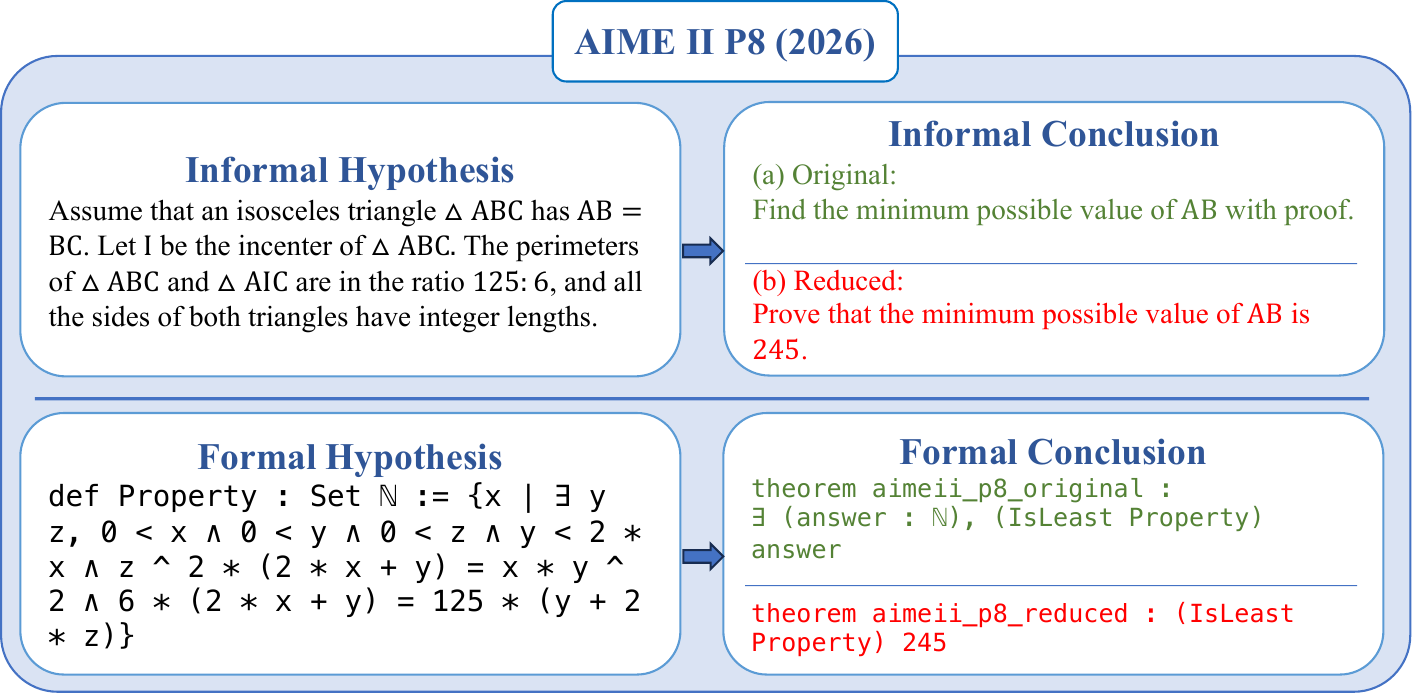}
    \caption{Example American Invitational Mathematics Examination (AIME) II Problem 8 (2026). \textbf{Top:} Informal problem statement in natural language, where the original task requires constructing an answer and proving it satisfies the statement, while the reduced task asks for a proof after the ground-truth answer has been substituted. \textbf{Bottom:} Formalized statement in Lean 4.}
    \label{fig:aimeiip8}
\end{figure}

We focus on math competition problems in Lean, which broadly fall into two types. \textit{Theorem-proving} problems have the form $\forall X.\,P(X)\rightarrow Q(X)$: the task is to prove a fully specified target statement. \textit{Answer-construction} problems have the form $\exists \Psi.\,\forall X.\,P(X)\rightarrow Q(X,\Psi(X))$: the answer object $\Psi$ is not supplied and must be synthesized together with a proof of correctness. Once $\Psi$ is known, an answer-construction problem reduces to the theorem-proving task of proving the substituted statement.
Figure~\ref{fig:aimeiip8} illustrates this distinction using AIME II (2026) Problem 8: the original problem asks for an answer and a proof, while the reduced version substitutes the ground-truth answer and asks only for a proof.

Existing methods and benchmarks in Lean mainly focus on theorem-proving problems. Popular benchmarks like ProofNet~\citep{proofnet}, MiniF2F~\citep{minif2f}, and FormalMath~\citep{formalmath} are formalized from informal source problems, and answer-construction problems in the original informal sources are reduced to theorem-proving problems by substituting the ground-truth answers. Benchmarks like PutnamBench~\citep{putnambench} include answer-construction problems, but these problems are reduced to theorem-proving versions in practice. In fact, according to the PutnamBench leaderboard for the Lean proof assistant\footnote{Source: \href{https://trishullab.github.io/PutnamBench/leaderboard.html}{https://trishullab.github.io/PutnamBench/leaderboard.html}.}, there are 33 ``with-answer'' (theorem-proving) evaluations, where the ground-truth answer is already substituted into the theorem statement, but only 2 ``no-answer'' (answer-construction) evaluations, where the answer needs to be constructed. Our method was ranked first in ``no-answer'' evaluation at the time of submission in June 2026. Current prover LLMs~\citep{deepseekprover,deepseekproverv15,deepseekproverv2,lin2025goedel,wang2025kimina} focus on theorem-proving problems rather than answer-construction problems, and pure symbolic search is expensive for various domains~\citep{collins1976quantifier,fu2023complexity}, with real quantifier elimination doubly exponential in the worst case~\citep{davenport1988real}.

\begin{figure}[ht!]
    \centering
    \includegraphics[width=1.0\linewidth]{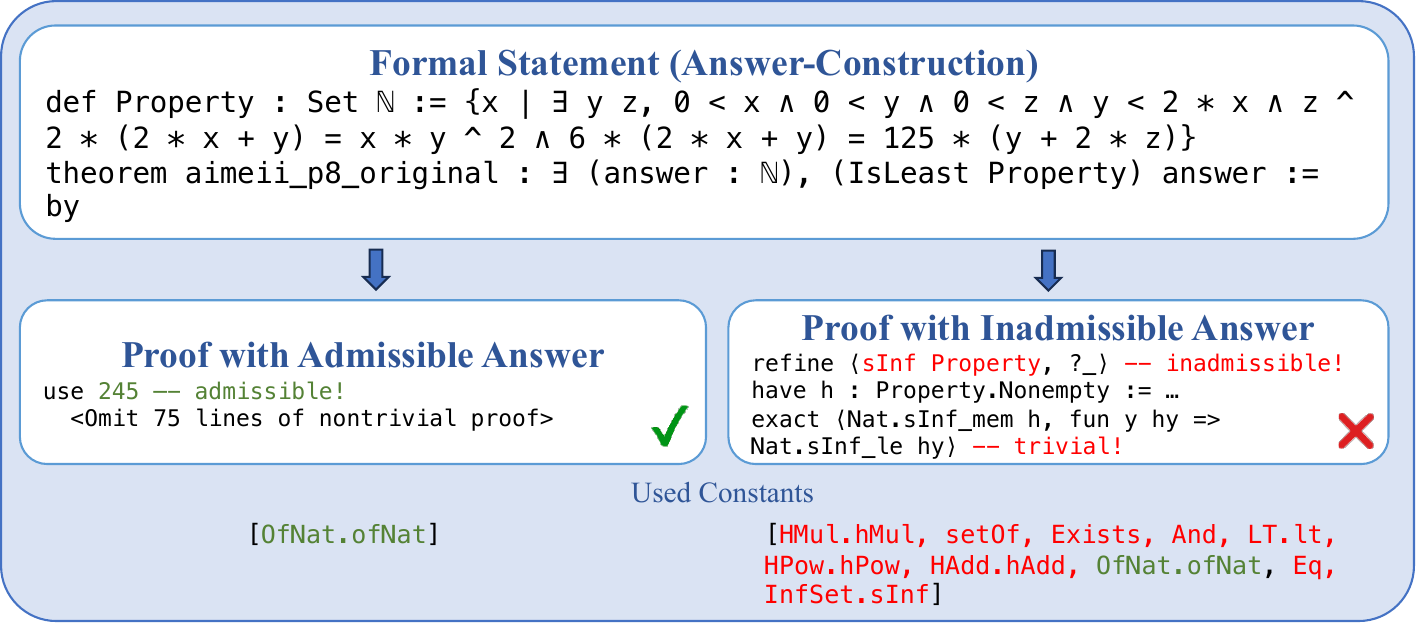}
    \caption{Two compiling Lean proofs for AIME II P8 (2026) with different constructed answers. \textbf{Left:} Proof with an admissible natural-number numeral. \textbf{Right:} Proof with an inadmissible circular answer. Complete proofs are in Appendices~\ref{app:aimeii8_successful_proof} and~\ref{app:aimeii8_inadmissible_proof}.}
    \label{fig:aimeiip8_inadmissible_example}
\end{figure}

The difficulty of formally solving answer-construction problems is twofold: (i) \textit{admissibility checking for constructed answers} and (ii) \textit{misaligned capabilities between general and prover LLMs}.
For (i), the issue is as follows: in an answer-construction problem, we require the constructed answer to have the target format, such as ``a natural-number numeral'' (e.g. $1$) or ``a real number with arithmetic expressions'' (e.g. $\frac{5}{2}$). We call such answers admissible answers.
However, Lean does not support checking whether a constructed answer is in the target format or simply ``cheats'' by circularly constructing a non-canonical answer that trivializes the proof. We show two proofs with different constructed answers in Figure~\ref{fig:aimeiip8_inadmissible_example}: the AIME problem expects the answer to be a natural-number numeral, and both displayed proofs are accepted by the Lean compiler. The first proof's answer is $245$ and is clearly admissible.
However, the second proof's answer \lstinline|sInf Property| essentially says ``the minimum possible value of $AB$ is the minimum possible value of $AB$''. Although the second proof passes the Lean check, it is clearly inadmissible and trivializes the proof via a circular argument.
In fact, as shown in our evaluation in Section~\ref{sec:empirical_eval}, simply using a general or prover LLM to solve answer-construction problems can produce many spurious ``successful'' trivial proofs that contain inadmissible answers.
For (ii), the issue is as follows: general LLMs are effective at constructing answers due to their strong informal reasoning capabilities and agentic tool calls, but are expensive; prover LLMs are fine-tuned on vast formal proof data and have higher formal proof-generation accuracy than general LLMs under aligned parameter sizes and cost budgets, but are mostly designed for pure theorem-proving tasks and are less reliable at answer construction.

To address the limitations of existing work, we propose Enumerate-Conjecture-Prove (\framework{}), an end-to-end framework for formally solving answer-construction problems. \framework{} first leverages general LLMs for enumeration and answer construction, then generates formal proofs via prover LLMs with answer refinement upon disproof.
\framework{} verifies the solution through admissibility checking for the constructed answer and Lean compilation of the proof. Overall, \framework{} takes advantage of general and prover LLMs to correctly solve more answer-construction problems at lower cost, while ensuring the admissibility of constructed answers to avoid trivializing proofs. The \framework{} workflow is largely inspired by P\'olya's problem-solving methodology~\citep{polya2014solve}, which emphasizes exploration of simpler cases, pattern generalization, and proof. The answer refinement loop is also connected to work on counterexample-guided abstraction refinement~\citep{satcegar,cegar}.

We organize the rest of the paper as follows. We first introduce preliminary notation and formulations of answer-construction problems in Section~\ref{sec:preliminaries}, followed by a discussion of related work in Section~\ref{sec:related_work}. We present our method, including problem preprocessing, answer construction, and proving, in Section~\ref{sec:methods}. We then describe the benchmarks and present the evaluation results in Section~\ref{sec:empirical_eval}. Finally, we summarize \framework{} with an analysis of limitations and future work in Section~\ref{sec:conclusion}.

\section{Preliminaries}
\label{sec:preliminaries}
In this section, we introduce mathematical notation and basic Lean background, then describe helper functions and formulations of answer-construction problems that illustrate the \framework{} workflow.

\subsection{Basic Notation}
By convention, we denote by $D$ the universe of discourse (domain) and by $V$ the set of variables, each denoted by a lowercase letter with optional subscripts.
Given sequences of variables $X=\langle x_1,\ldots,x_n\rangle$ and $Y=\langle y_1,\ldots,y_m\rangle$, and a predicate $P$ over $X$ and $Y$, we write $\forall X. \exists Y. P(X,Y)$ as shorthand for $\forall x_1. \cdots \forall x_n. \exists y_1. \cdots \exists y_m. P(x_1,\ldots,x_n,y_1,\ldots,y_m)$. When the variables are clear from the context, we write $P$ as shorthand for $P(X,Y)$, and $\text{Vars}(P)$ for the set of variables occurring free in $P$.
An assignment to a formula $P$ is a mapping $\sigma: \text{Vars}(P) \to D$ that assigns each variable a concrete value in the domain; $\sigma$ satisfies $P(X,Y)$ if $P$ evaluates to $\mathrm{True}$ under $\sigma$. For a variable $y_i \in Y$ and a value $b \in D$, we write $P(X,Y)_{|y_i=b}$ for the formula obtained by substituting all free occurrences of $y_i$ in $P$ with $b$.
For a tuple $Y'=\langle y_1',\ldots,y_m'\rangle$, we write $P(X,Y')$ for the simultaneous substitution of each $y_i$ with $y_i'$ in $P$.
We say that a set is \emph{extensionally defined} if it is an explicit finite enumeration of elements (e.g., $\{1,2,3\}$), or \emph{intensionally defined} if it is specified by a predicate that characterizes the elements belonging to the set (e.g., $\{x \in \mathbb{R} \mid x > 0\}$).

\subsection{Lean 4 and Dependent Type Theory}
\label{sec:dependent_type_theory}
Lean 4~\citep{lean} is a proof assistant and functional programming language whose kernel implements a dependent type theory in the family of the Calculus of Inductive Constructions (CIC). Mathlib~\citep{mathlib} is the largest Lean standard library of definitions, theorems, and proofs in cutting-edge formalized mathematics.
In Lean, propositions are types and proofs are terms according to the Curry-Howard correspondence.
A \emph{type judgment} has the form $\Gamma \vdash t : T$, meaning that the term $t$ has type $T$ under a typing context $\Gamma$ (a list of variable bindings).
The language is based on a small core of term formers: variables, constants (definitions) from the Lean environment $\Sigma$, function application, $\lambda$-abstraction, dependent function types (also written as $\forall$-types), and let-bindings. Example constants include \texttt{Real.pi} (denoting $\pi$) and functions such as \texttt{Real.sin} (denoting $\sin(\cdot)$) and \texttt{HAdd.hAdd} (denoting $+$).
The Lean elaborator translates Lean code into this kernel representation by inserting \emph{implicit} arguments, such as universe parameters and type arguments, and \emph{instance} arguments, such as typeclass dictionaries.
To align our presentation with standard dependent type theory, we describe Lean expressions using a compact core calculus according to LeanAuto's~\citep{leanauto} $\lambda C$ type system with a countable number of non-cumulative universe levels. Let $U_\ell$ ($\ell \in \mathbb{N}$) denote the sorts, let $V$ denote the set of variables, and let $\Sigma$ denote the set of constants.
The syntax of expressions is inductively defined as follows:
\begin{align}
T ::= {} & V \mid \Sigma \mid U_\ell \mid T\,T
    \mid \lambda (V : T).\,T \notag  {}\mid \forall (V : T).\,T
    \mid \text{let}\ (V : T) := T;\,T.
\end{align}
Here, $T\,T$ denotes function application, $\lambda (V : T).\,T$ denotes $\lambda$-abstraction, $\forall (V : T).\,T$ denotes a dependent product type, and $\text{let}\ (V : T)$ denotes a let-binding.
For example, \lstinline|Real.sin x| corresponds to $T\,T$ with head constant \lstinline|Real.sin| and variable $x$, and \lstinline|HAdd.hAdd a b| corresponds to $((\texttt{HAdd.hAdd}\ a)\ b)$, i.e., nested applications using $T\,T$. Similarly, \lstinline|fun x : Real => Real.sin x + 1| corresponds to $\lambda (x:\mathbb{R}).\,((+)\ (\sin\ x))\ 1$ after elaboration, where $+$ is a constant such as \lstinline|HAdd.hAdd| with implicit and instance arguments inserted.

\subsection{Helper Functions}
We use the following helper functions for notational convenience in algorithm descriptions. Let $\bot$ denote the empty string.
$\mathsf{LLM}(\textsf{prompt})$ queries a language model with an input prompt string $\textsf{prompt}$ and outputs a response string. $\mathsf{Python}(\mathsf{code})$ executes a Python program string $\mathsf{code}$ and returns its standard output and standard error. $\mathsf{Lean}(\mathsf{code})$ compiles a Lean program string $\mathsf{code}$ and returns $\mathsf{error\_message}$. When $\mathsf{code}$ is composed of a theorem statement and proof, $\mathsf{error\_message}=\bot$ indicates that the proof is correctly verified. We refer to \emph{general LLMs} as models trained for broad tasks, such as GPT-5.4~\citep{singh2025openai} and Qwen3~\citep{qwen3}, and to \emph{prover LLMs} as models fine-tuned for formal theorem proving in Lean, such as Goedel-Prover-V2~\citep{lin2025goedel}.

\subsection{Answer-Construction Formulations}
\label{sec:problem_formulation}
We model the \emph{answer-construction} problem as a \emph{functional-synthesis} problem: synthesize a mathematical object (an answer function) satisfying given specifications, together with a machine-checkable proof of correctness. Let $D$ be the universe of discourse, and let $V = X \cup Y$ be the variables, including \emph{context variables} $X = \langle x_1,\ldots,x_n\rangle$ and \emph{answer variables} $Y = \langle y_1,\ldots,y_m\rangle$.
The \emph{hypothesis predicate} $P(X)$ denotes the constraint for context variables, and the \emph{conclusion predicate} $Q(X,Y)$ denotes the constraint for both context and answer variables.
Given any $X$ satisfying $P(X)$, the answer function $\Psi(X)=\langle \psi_1(X),\ldots,\psi_m(X)\rangle$ must map each such $X$ to either a witness $Y$ satisfying $Q(X,Y)$ (\textit{find-any problem}) or the set of all witnesses $Y$ satisfying $Q(X,Y)$ (\textit{find-all problem}). In particular, a find-any problem is written as \(\exists \Psi.\forall X. P(X) \rightarrow Q(X,\Psi(X))\), and a find-all problem is written as \(\exists \Psi.\forall X. P(X) \rightarrow \forall Y\, (Y\in \Psi(X) \leftrightarrow Q(X,Y))\). The admissible vocabulary $A$ is the set of allowed constants that can appear in the expression of constructed answers, and the answer $\Psi$ is said to be \textit{admissible} if all constants it uses are in $A$. Overall, each answer-construction problem is specified by the 5-tuple $T=(D,V,P,Q,A)$. When the context is clear, we use $T$ to denote the formalized Lean statement for the answer-construction task, and $T_\Psi$ to denote the statement after the $\exists$ is eliminated by substituting the explicit $\Psi$. We say that a proof successfully solves the answer-construction problem $T$ if it compiles without error in Lean and the constructed answer in the proof is admissible.

\begin{example}
The AIME II P8 example in Figure~\ref{fig:aimeiip8} is a find-any problem. In fact, it does not have any free context variables, so $X:=\varnothing$ and $P := \text{True}$. The only answer variable is $Y=\langle y_1\rangle$ in domain $D=\mathbb{N}$, which is essentially the minimum possible value of the edge length $AB$ in the problem.
Notice that this problem can equivalently be stated as ``\textit{find the least positive integer \(x\) such that there exist positive integers \(y,z\) with \(y<2x\), \(z^2(2x+y)=xy^2\), and \(6(2x+y)=125(y+2z)\)}'' by translating the geometric information into algebraic form. Thus,
$Q(y_1) := \Bigl(y_1=\min\{y\in \mathbb{N}: \exists z_1,z_2\in \mathbb{N},(y,z_1,z_2>0)\wedge 2y>z_1\wedge\ z_2^2(2y+z_1)=yz_1^2\wedge\ 6(2y+z_1)=125(z_1+2z_2)\}\Bigr)$. The formalized statement is obtained from this information.
Since the AIME competition requires the synthesized $\Psi$ to be a canonical natural number without arithmetic operations, the admissible vocabulary is $A=$\texttt{[OfNat.ofNat]}.
\end{example}

\section{Related Work}
\label{sec:related_work}
\paragraph{Prover LLMs in Lean.}
Interactive theorem provers (ITPs) such as Isabelle~\citep{isabelle} and Lean~\citep{lean} support human-friendly environments for proof development. Lean
supports built-in proof automation tactics such as LeanSMT~\citep{leansmt} and aesop~\citep{limperg2023aesop}. Recent prover LLMs combine these formal environments with large-scale proof search and training on a vast formal corpus. Open-weight prover LLMs such as Goedel-Prover~\citep{lin2025goedel}, DeepSeek-Prover~\citep{deepseekprover,deepseekproverv15,deepseekproverv2}, InternLM-Prover~\citep{internlm}, and Kimina-Prover~\citep{wang2025kimina} substantially improve Lean proof success on standard Lean theorem-proving benchmarks such as miniF2F and PutnamBench. Inference-time techniques like Seed-Prover~\citep{chen2025seed} and Aristotle~\citep{aristotle} maintain lemma pools of proved conjectures to optimize the efficacy of prover LLMs and achieve silver- to gold-medal-level results in IMO.
These systems are primarily evaluated on theorem-proving (non-answer-construction) problems with specified conclusions, whereas our answer-construction setting focuses on tasks with existentially quantified answer variables.

\paragraph{Answer-Construction Systems.}
Several systems address answer-construction problems. PyEuclid~\citep{pyeuclid} combines an AlphaGeometry-style deduction engine~\citep{alphageometry} with algebraic computation to solve Euclidean geometry calculation problems with proofs in a geometry DSL. FPS~\citep{liu2025beyond} studies general answer-construction problems in Lean 4 through a search-based algorithm, and reports solving 1 PutnamBench problem in the answer-construction format. High-performance formal reasoning engines have also shown strong results on Olympiad-level problems. AlphaProof~\citep{alphaproof} solved three out of six IMO 2024 problems in Lean 4 and one more problem in the AlphaGeometry DSL, contributing to silver-medal performance. It also leverages general LLMs for answer-construction problems to sample hundreds of candidate answers and attempt proofs on the substituted Lean statements. However, AlphaProof's model weights and implementation are closed-source. Moreover, its training and inference pipelines are substantially more expensive: in fact, AlphaProof reports around 180,000 TPU-days of Google v6e for data processing and reinforcement learning, several orders of magnitude more compute than our experiments use.

\paragraph{Formal Benchmarks.}

General formal benchmarks such as MiniF2F~\citep{minif2f}, FIMO~\citep{fimo}, LeanWorkBook~\citep{workbook}, ProofNet~\citep{proofnet}, and FormalMath~\citep{formalmath}, as well as domain-specific benchmarks like LeanGeo~\citep{song2025leangeo} (geometry), MO-INT~\citep{wei2024proving} (inequality), and FormalML~\citep{formalml} (machine learning theory), are constructed by formalizing raw informal mathematics from math textbooks and past competition problems. These benchmarks focus on formal theorem-proving problems, and the ground-truth answers are substituted into the statements during formalization if the raw informal statements are answer-construction problems. PutnamBench~\citep{putnambench} (Putnam competition) and CombiBench~\citep{combibench} (combinatorics) include some answer-construction problems. However, most methods still evaluate on the reduced theorem-proving versions of problems, and there is no formal mechanism to detect the admissibility of candidate constructed answers.

\section{Methods}
\label{sec:methods}
\lstset{style=colored}

\begin{figure}[ht!]
    \centering
    \includegraphics[width=0.6\linewidth]{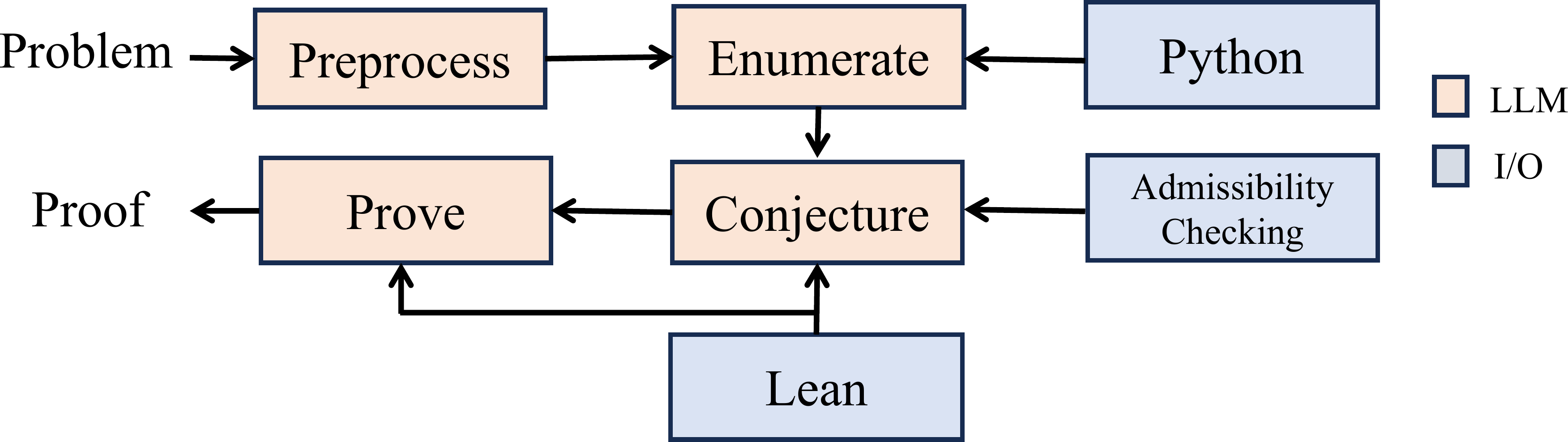}
    \caption{An overview of the \framework{} workflow.}
    \label{fig:ecp}
\end{figure}

In this section, we describe the main workflow of \framework{}. Prior to problem solving, we first preprocess the raw answer-construction problems into Lean formalizations with admissible vocabularies according to the template in Section~\ref{sec:problem_formulation}. Then \framework{} solves the answer-construction problem as follows. (i) During
the Enumerate and Conjecture stages, the general LLM constructs an answer by first performing bounded Python enumeration and then generalizing to the broader case to conjecture an admissible answer in Lean. (ii) During the Prove stage, the constructed answer is substituted into the theorem statement and the prover LLM generates the proof with automation tactics. \framework{} returns to the Enumerate and Conjecture stage if the substituted statement can be disproved. An overview of \framework{} is shown in Figure~\ref{fig:ecp}. We explain each stage in detail in the subsequent subsections.

\subsection{Preprocess Stage}
\label{sec:preprocessing}
\paragraph{Autoformalization.}
During the Preprocess stage, the raw informal problem statement in natural language is converted into a formal problem with an admissible vocabulary according to the problem formulation template in Section~\ref{sec:problem_formulation}. Specifically, we prompt a general LLM with the raw informal problem and instructions for producing the problem formulation according to the template in Appendix~\ref{app:prompts}, and ask it for a compilable formalization. Each formalization is validated using Lean compilation for syntax correctness and manual inspection for equivalence to its informal statement. Failed formalizations are sent to the LLM for repair. Using this approach, we autoformalized the 75 problems in the 2026 MathArena Final Answer Competitions~\citep{matharena}, consisting of 30 AIME problems, 33 HMMT (Harvard-MIT Mathematics Tournament) problems, and 12 Apex problems. In addition, out of 672 existing formalized problems in PutnamBench~\citep{putnambench}, we found that 346 are answer-construction problems and processed them into our problem formulation.
\paragraph{Admissible Vocabularies.} We label the admissible vocabularies for each problem using rule-based heuristics according to each problem's source and answer type. In particular, we design three levels of vocabulary breadth for different problem sources: (i) basic numbers without arithmetic operations (for AIME problems in MathArena); (ii) expressions with elementary arithmetic operations such as $+$, $-$, $\times$, and $\div$ (for HMMT and Apex problems in MathArena); (iii) expressions with university-level constants (for PutnamBench problems). We also infer vocabularies from answer types: if the expected answer has tuple-, set-, or function-related constructor constants, detected by occurrences of \texttt{\texttimes{}}, \texttt{Set}, or \texttt{\textrightarrow{}} in the answer type, we include additional vocabulary accordingly.

After labeling, we run a validation test to examine whether each problem's labeled admissible vocabulary covers the constants actually used in the ground-truth answer. We found that
all 75/75 MathArena problems and
301/346 (87.0\%) PutnamBench problems passed validation. We discovered that the remaining 45/346 (13.0\%) PutnamBench problems have complicated answer formats and manually fixed them as described in Appendix~\ref{app:admissible_vocabularies}.

\subsection{Admissibility Checking}

In addition to semantic correctness, answer-construction problems in Lean require that the synthesized answer be admissible, meaning that it is composed of constants in the admissible vocabulary. For example, in the AIME II P8 example in Figure~\ref{fig:aimeiip8}, answers like \texttt{1} or \texttt{245} are admissible, but answers like \texttt{sInf Property} are inadmissible. To check admissibility, it is sufficient to extract the constants used by the answer $\Psi$ following Lean's expression syntax from Section~\ref{sec:dependent_type_theory}. Let $\Sigma$ be the set of global constants, let $A \subseteq \Sigma$ be the admissible vocabulary, and let $A_{\mathsf{admin}}\subseteq\Sigma$ denote the set of \emph{administrative} constants that arise from the Lean compiler runtime, including type formers, typeclass instances, instance-producing constants, and compiler-generated private constants. We ignore constants in $A_{\mathsf{admin}}$ during admissibility checking.

For an application expression $t$, we use spine decomposition:
\(
t = h\ a_1\ a_2\ \cdots\ a_n,
\)
where $h$ is the \emph{head} and $a_1,\ldots,a_n$ are the arguments. We define $\mathsf{ExplicitArgs}(t)$ using Lean's elaborated function-argument metadata for explicit binders. In the projection case, $\pi_i$ denotes the corresponding field projection constant.
The constants used by the expression $t$, $\mathsf{UsedConstants}(t)$, are inductively defined as follows:
\begin{equation}
\label{admissible}
\mathsf{UsedConstants}(t)=
\begin{cases}
\{c\} & t=c,\ c\in\Sigma\setminus A_{\mathsf{admin}} \\

\displaystyle\bigcup_{t'\in\{A,b\}}\mathsf{UsedConstants}(t')
  & t=\lambda(x:A).\, b \\
\displaystyle\bigcup_{t'\in\{A,B\}}\mathsf{UsedConstants}(t')
  & t=\forall(x:A).\, B \\
\displaystyle\bigcup_{t'\in\{A,u,v\}}\mathsf{UsedConstants}(t')
  & t=\mathbf{let}\ (x:A):=u;\, v \\
\mathsf{UsedConstants}(t')\cup \{\pi_i\}
  & t=\pi_i(t') \\
\displaystyle\bigcup_{t'\in\{h\}\cup\mathsf{ExplicitArgs}(t)}\mathsf{UsedConstants}(t')
  & t=h\,a_1 \cdots a_n \\
\emptyset & \text{otherwise}
\end{cases}
\end{equation}

We define admissibility of the constructed answer $\Psi$ with respect to $A$, $\mathsf{Admissible}(A, \Psi)$, as
$\mathsf{UsedConstants}(\Psi)\subseteq A$. Existential and universal quantifiers require slightly different handling. First, $\exists$ can be treated as the constant \lstinline|Exists| because the expression $\exists (x : A). B$ is elaborated as \lstinline|Exists (fun x: A => B)| in Lean. This expression is handled by the application case $t=h\,a_1\cdots a_n$, with \lstinline|Exists| as the head $h$, together with the lambda-abstraction case $t=\lambda (x:A).b$ for \lstinline|(fun x: A => B)|. Second, if $\forall$ is not allowed, then the expression $t$ is inadmissible whenever the case $t=\forall (x:A).B$ is executed during $\mathsf{UsedConstants}(t)$.

\subsection{Enumerate and Conjecture Stage}
\label{sec:enumerate}
During the Enumerate and Conjecture stage, \framework{} is given the Lean answer-construction problem statement $T$ and the expected answer type. It first writes Python programs for bounded enumeration (Enumerate stage), then constructs an answer that generalizes from enumeration with mathematical reasoning (Conjecture stage), as described in Algorithm~\ref{alg:constructanswer}.

\paragraph{LLMs with Tool Calling.}
General LLMs, such as the commercial model GPT-5.4 and the open-source local model Qwen3, support reasoning with tool calls for code generation and execution. In particular, the LLM encloses tool calls (code content) inside delimiters \texttt{<tool\_call>\ldots</tool\_call>} under the OpenAI function-calling JSON schema. The compiler backend executes the tool call and adds stdout and stderr to the rollout response using delimiters \texttt{<tool\_response>\ldots</tool\_response>} immediately after \texttt{</tool\_call>}. For fast and parallel code execution, we deploy the SandboxFusion~\citep{sandboxfusion} server as the code interpreter backend, and we restrict the tool interface to Python calls. Let \(\mathsf{ParseToolCall}\) denote the helper function that extracts the tool call from a response.

\paragraph{Enumerate Stage.}
Initially, the prompt is built from the system instruction template in Appendix~\ref{app:prompts} and the input examples.
\framework{} first queries a general LLM to write one or more Python programs to directly solve for the answer or search for tuples $(X,Y)$ satisfying $P(X)\wedge Q(X,Y)$ over bounded and tractable values, as shown in Lines 2--7 in Algorithm~\ref{alg:constructanswer}. Each Python program is executed by SandboxFusion, and each program and its output are added back to the context. There may be multiple rounds of tool calls, depending on whether the enumeration results have provided sufficient evidence for answer construction. When the LLM no longer generates tool calls or hits $\mathsf{max\_enumerate\_rounds}$, it proceeds to the Conjecture stage.

\begin{algorithm}[ht!]
\caption{$\mathsf{ConstructAnswer}(T, \mathsf{FailedAnswers}=\bot)$}
\label{alg:constructanswer}
\begin{algorithmic}[1]
\State $\mathsf{context} \gets \mathsf{BuildPrompt}(T,\mathsf{FailedAnswers})$

\For{$i \gets 1$ \textbf{to} $\mathsf{max\_enumerate\_rounds}$}
    \State $\mathsf{response} \gets \mathsf{LLM}(\mathsf{context})$
    \State $\mathsf{code} \gets \mathsf{ParseToolCall}(\mathsf{response})$
    \If{$\mathsf{code} = \bot$}
        \State \textbf{break}
    \EndIf
    \State $\mathsf{context} \gets \mathsf{context} + \mathsf{code} + \mathsf{Python}(\mathsf{code})$
\EndFor
\For{$i \gets 1$ \textbf{to} $\mathsf{max\_conjecture\_rounds}$}
    \State $\mathsf{response} \gets \mathsf{LLM}(\mathsf{context})$
    \State $\Psi \gets \mathsf{ExtractAnswer}(\mathsf{response})$
    \State $ \mathsf{error\_message} \gets \mathsf{Lean}(\Psi)$
    \State $\mathsf{is\_admissible}\gets \mathsf{Admissible}(A, \Psi)$
    \If{$\mathsf{error\_message}=\bot$ \textbf{and} $\mathsf{is\_admissible}$}
        \State \Return $\Psi$
    \EndIf
    \If{$\mathsf{error\_message}\neq\bot$}
        \State $\mathsf{context}\gets \mathsf{context}+\Psi+\mathsf{error\_message}$
    \EndIf
    \If{\textbf{not} $\mathsf{is\_admissible}$}
        \State $\mathsf{context}\gets \mathsf{context}+\Psi+\text{``inadmissible''}$
    \EndIf
\EndFor
\State \Return $\mathsf{\bot}$
\end{algorithmic}
\end{algorithm}

\begin{figure}[ht!]
    \centering
    \includegraphics[width=1.0\linewidth]{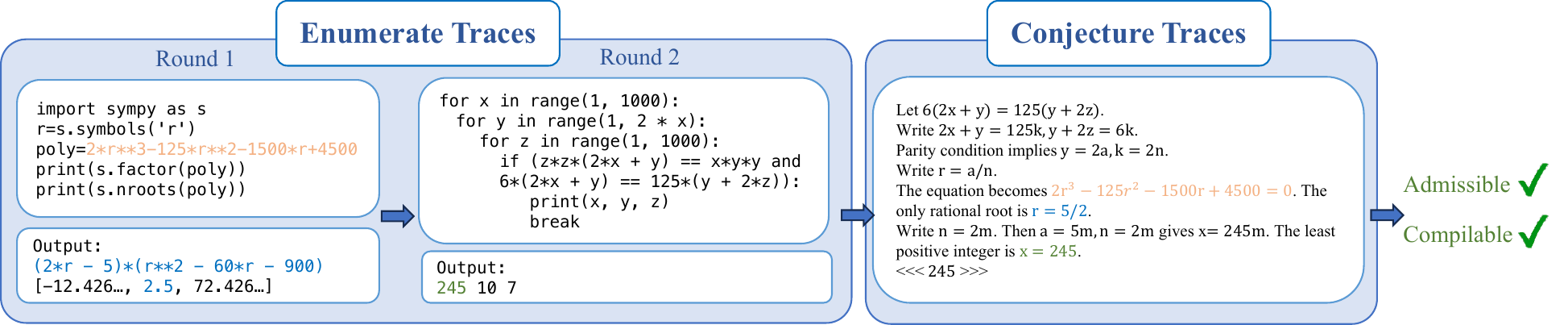}
    \caption{LLM traces for solving AIME II P8 (2026) during the Enumerate (\textbf{Left}) and Conjecture (\textbf{Right}) stages. The Conjecture trace is manually compacted due to the page limit.}
    \label{fig:aimeiip8_traces}
\end{figure}

\paragraph{Conjecture Stage.}
During the Conjecture stage, the LLM sees the Python programs and execution results produced by the Enumerate stage, performs mathematical reasoning toward the answer construction by generalizing beyond bounded enumerations, and proposes the explicit final answer $\Psi$ inside the delimiter \texttt{<}\texttt{<}\texttt{<}\ldots\texttt{>}\texttt{>}\texttt{>} for extraction (denoted by $\mathsf{ExtractAnswer}(\mathsf{response})$). \framework{} then runs Lean compilation and an admissibility check for the constructed answer, and returns the answer for repair if it fails either test.

An explicit example of LLM traces during answer construction for AIME II P8 (2026) is shown in Figure~\ref{fig:aimeiip8_traces}. During the Enumerate stage, the LLM first uses SymPy to solve for a key equation with rational root $r=\frac{5}{2}$ and writes a brute-force enumeration to print the smallest valid answer. In the Conjecture stage, it leverages the key enumerated results, performs mathematical reasoning, and concludes that the answer is $245$, which compiles in Lean and is admissible.

\subsection{Prove Stage}
\label{sec:prove}
After the Conjecture stage, the constructed answer is substituted into the theorem statement to form $T_\Psi$ using the Lean tactic \lstinline{use Ψ}. The goal of the Prove stage is to find a proof of $T_\Psi$.
It first applies Lean's automation engines, denoted by $\mathsf{Auto}$, by sequentially applying the automation tactics \texttt{native\_decide}, \texttt{simp}, \texttt{aesop}~\citep{limperg2023aesop}, \texttt{ring}, and \texttt{norm\_num}. If automation fails, it invokes a prover LLM with a repair loop that incorporates error messages from unsuccessful attempts, as shown in Algorithm~\ref{alg:prove_simplified}.
In practice, this procedure runs for multiple parallel independent attempts and succeeds if any attempt outputs a successful proof.

In \framework{}, $\mathsf{Prove}$ is invoked for both $T_\Psi$ and its negation $\neg T_\Psi$.
A successful $\mathsf{Prove}(T_\Psi)$ concludes the \framework{} loop; if $\mathsf{Prove}(\neg T_\Psi)$ succeeds, \framework{} adds $\Psi$ to the failed answer collection and returns to the Enumerate and Conjecture stage for answer reconstruction. We leave the full proof of the AIME II P8 example in Figure~\ref{fig:aimeiip8} to Appendix~\ref{app:aimeii8_successful_proof}.

\begin{algorithm}[ht!]
\caption{$\mathsf{Prove}(T_\Psi)$}
\label{alg:prove_simplified}
\begin{algorithmic}[1]
\For{$j\gets 1$ \textbf{to} $\mathsf{parallel\_proof\_samples}$}
    \State $\mathsf{feedback} \gets\bot$

    \State $ \mathsf{error\_message} \gets \mathsf{Lean}({T_\Psi} + \mathsf{automation})$
    \If{$\mathsf{error\_message}=\bot$}
        \State \Return $\mathsf{automation}$
    \EndIf
    \For{$i \gets 1$ \textbf{to} $\mathsf{max\_prove\_rounds}$}

        \State $\mathsf{proof}\gets \mathsf{LLM}({T_\Psi} +\mathsf{feedback})$
        \State $ \mathsf{error\_message} \gets \mathsf{Lean}(\mathsf{proof})$
        \If{$\mathsf{error\_message}=\bot$}
            \State \Return $\mathsf{proof}$
        \EndIf
        \State $\mathsf{feedback}\gets \mathsf{proof}+\mathsf{error\_message}$
    \EndFor
\EndFor
\State \Return $\bot$
\end{algorithmic}
\end{algorithm}

\begin{algorithm}[ht!]
\caption{$\mathsf{ECP}(T)$}
\label{alg:ecp_loop}
\begin{algorithmic}[1]
\State $\mathsf{FailedAnswers} \gets \bot$

\For{$r \gets 1$ \textbf{to} $\mathsf{max\_ecp\_rounds}$}
    \State $\Psi\gets\mathsf{ConstructAnswer}(T, \mathsf{FailedAnswers})$
    \If{$\Psi= \bot $}
        \State \Return $\bot$
    \EndIf
    \State $\textsf{proof}\gets \mathsf{Prove}(T_{\Psi})$
    \State $\textsf{disproof}\gets  \mathsf{Prove}(\neg T_{\Psi})$
    \If{$\textsf{proof}\neq \bot$}
        \State \Return \texttt{use }$\Psi$\texttt{;} + $\mathsf{proof}$
    \EndIf
    \If{$\mathsf{disproof}\neq \bot$}
        \State $\mathsf{FailedAnswers}\gets \mathsf{FailedAnswers} + \Psi$
        \State \textbf{continue}
    \EndIf
    \State \textbf{break}
\EndFor
\State \Return $\bot$
\end{algorithmic}
\end{algorithm}

\subsection{Equivalence Testing for Answers}
\label{sec:equality_testing}
In addition to end-to-end proof generation, this approach naturally applies to equivalence checking between the ground-truth answer \(\Psi_{\text{ground\_truth}}\) and a constructed answer \(\Psi\) by proving the Lean goal \(\Psi = \Psi_{\text{ground\_truth}}\), when the problem has a find-all type or a find-any type with a unique answer. This test uses the fixed tactic portfolio and prover LLMs described in Section~\ref{sec:prove}, making the check type-aware and certified whenever it succeeds. It can also prove equivalences that may be missed by script-based symbolic checkers, such as MathArena's evaluation protocol~\citep{matharena}, or by traditional LLM judges.

\section{Evaluation}
\label{sec:empirical_eval}
In this section, we evaluate \framework{} against baseline methods for solving answer-construction problems across multiple math competition benchmarks in Lean. The evaluated benchmarks span difficulty levels from high school to university and
include contamination-free problems beyond the evaluated models' knowledge cutoffs. Our evaluation aims to answer the following research questions (\textbf{RQs}):

\begin{enumerate}[label=\textbf{RQ \arabic*.}, leftmargin=*]
\item \textbf{Problem-Solving Accuracy.} Can \framework{} successfully solve more answer-construction problems in Lean than baseline methods?

\item \textbf{Answer-Only Accuracy.} Can \framework{} predict more correct answers without proof for answer-construction problems than baseline methods?

\item \textbf{Cost Efficiency.} Can \framework{} solve formal answer-construction problems at a lower cost than baseline methods?

\item \textbf{Ablation Studies.} What are the contributions of each component in \framework{}?

\end{enumerate}

\paragraph{Dataset Setup.}

\begin{figure}[H]
    \centering
    \begin{subfigure}[t]{0.4\columnwidth}
        \centering
        \includegraphics[width=\linewidth]{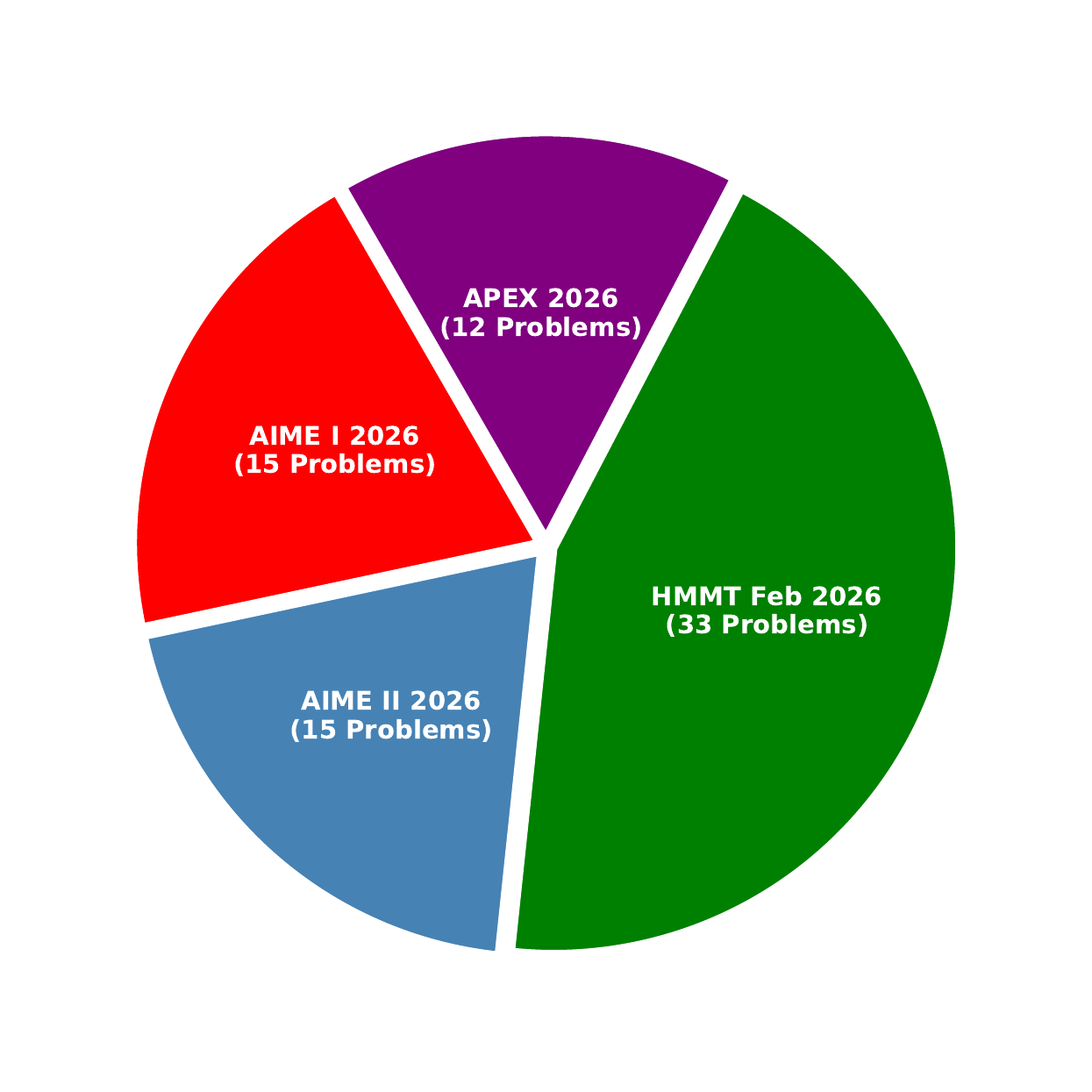}
        \caption{Sources}
        \label{fig:problem_sources}
    \end{subfigure}
    \hfill
    \begin{subfigure}[t]{0.4\columnwidth}
        \centering
        \includegraphics[width=\linewidth]{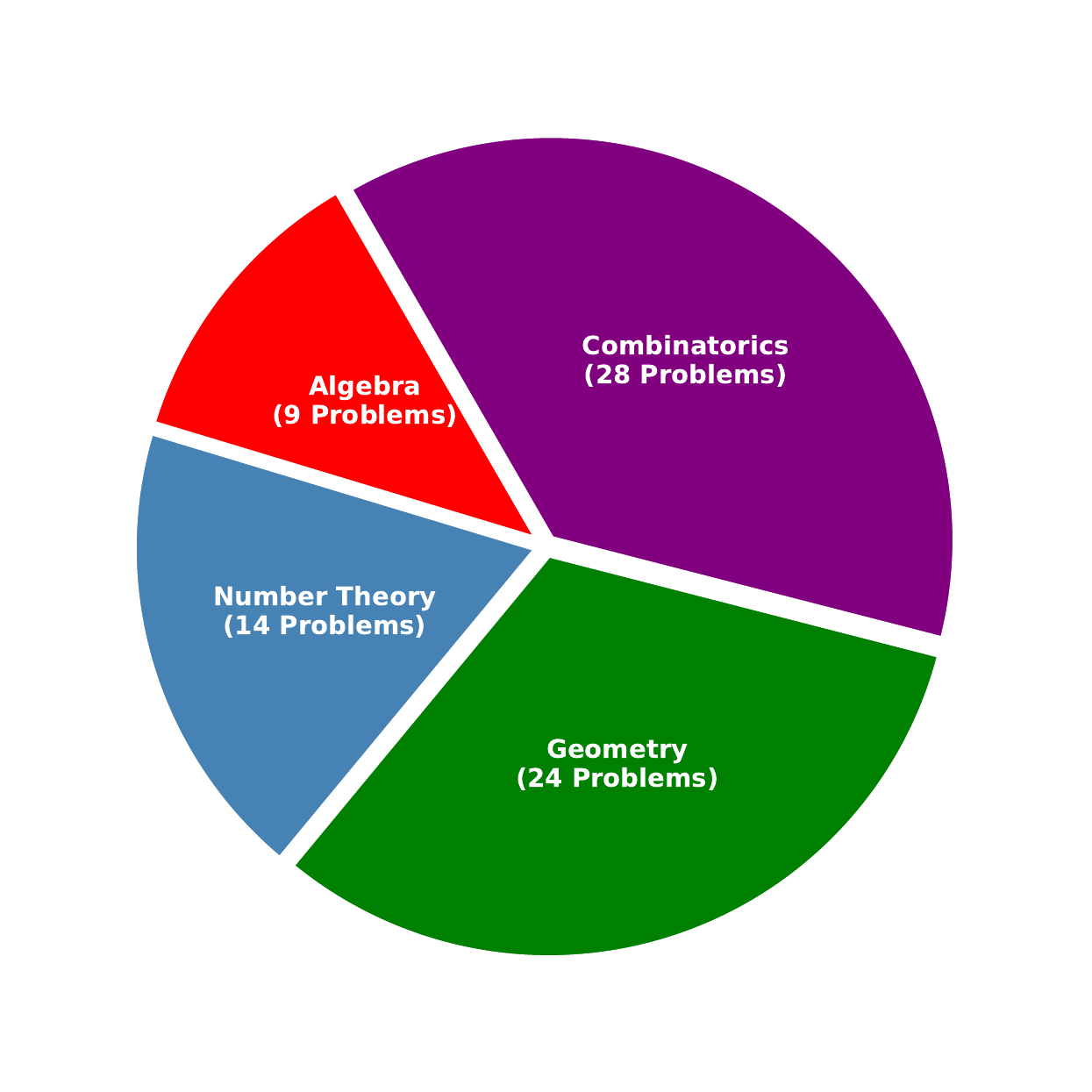}
        \caption{Domains}
        \label{fig:problem_categories}
    \end{subfigure}
    \caption{Problem sources and domains for MathArena 2026.}
    \label{fig:problem_sources_categories}
\end{figure}

We evaluate on two benchmarks: PutnamBench~\citep{putnambench} and MathArena~\citep{matharena}. PutnamBench originally includes 672 Putnam Mathematical Competition problems from 1962--2025 that are manually formalized in Lean 4 by human experts. We extract the subset of 346 instances that are answer-construction problems. The MathArena Final Answer Competitions benchmark\footnote{Source: \href{https://matharena.ai/?comp=overall--final_answer_comps&view=problem}{https://matharena.ai/?comp=overall--final\_answer\_comps\&view=problem}}~\citep{matharena} contains 75 problems from the AIME I, AIME II, HMMT (February), and Apex competitions in 2026, all of which are contamination-free for the evaluated models according to Table~\ref{tab:cutoff}. Since the raw MathArena problems are originally written in natural language, we preprocess them into the target formulation of answer-construction problems following Section~\ref{sec:preprocessing}. Figure~\ref{fig:problem_sources_categories} summarizes the MathArena sources and domains.

\begin{wraptable}{r}{0.4\textwidth}
\centering
\begingroup
\small
\setlength{\tabcolsep}{2pt}
\begin{tabular}{@{}lc@{}}

\toprule
\textbf{Dataset / Model} & \textbf{Knowledge Cutoff} \\
\midrule
PutnamBench~\citep{putnambench} & 1962--2025 \\
MathArena~\citep{matharena} & After Feb 2026 \\
\midrule
GPT-5.4~\citep{singh2025openai} & Before Aug 31, 2025 \\
Qwen3-32B~\citep{qwen3} & Before Mar 31, 2025 \\
Goedel-Prover-V2~\citep{lin2025goedel} & Before Aug 5, 2025 \\
\bottomrule

\end{tabular}
\endgroup
\caption{Knowledge cutoffs of evaluated datasets and models.}
\label{tab:cutoff}
\end{wraptable}

\paragraph{Hyperparameter Setup.}
We evaluate the general LLMs GPT-5.4 and Qwen3-32B, and the prover LLM Goedel-Prover-V2-32B.
For the \framework{} setup, we set the Enumerate, Conjecture, and ECP round limits to 10, 5, and 2, respectively. For the Prove stage, we use 4 and 32 parallel proof samples and a maximum of 3 prove rounds. Lean compilation is conducted on 64 AMD EPYC 9655 CPU cores in parallel with a 120-second timeout. The Qwen3 and Goedel-Prover-V2-32B models are deployed on 4 H100 SXM (80GB) GPUs in parallel. The GPT-5.4 model is queried via the OpenAI API platform. The default model setup of \framework{} uses GPT-5.4 for the Enumerate and Conjecture stages, and Goedel-Prover-V2-32B for the Prove stage.

\paragraph{Metric Setup.}
For \textbf{RQ 1}, we evaluate \framework{}'s and the baseline models' accuracy for answer-construction problem solving. A proof is said to successfully solve problem $T$ if it compiles in Lean without error and the constructed answer in the proof is admissible, as described in Section~\ref{sec:preliminaries}. Recall that $\mathsf{parallel\_proof\_samples}$ in \framework{}'s $\mathsf{Prove}$ is aligned to the Pass@$k$ accuracy metric, which generates $k$ independent parallel outputs and succeeds if any proof is successful. We also compare baseline LLMs GPT-5.4 and Goedel-Prover-V2-32B without \framework{} under the same Pass@$k$ and proof-correction attempt settings, meaning that they have the same $\mathsf{parallel\_proof\_samples}$, $\mathsf{max\_prove\_rounds}$, and access to the proof automation portfolio. For \textbf{RQ 2}, we focus on answer prediction without proof. That is, we only evaluate the number of constructed answers that are provably equivalent to the ground-truth answers of each dataset. For \textbf{RQ 3}, we report the API costs and GPU hours of the methods. The OpenAI API rates of GPT-5.4 are 2.5 USD per 1M input tokens and 15.0 USD per 1M output tokens. Our GPUs are on computing clusters, but we assume the H100 GPU hourly rate is 2.0 USD\footnote{Source: Vast AI's H100 pricing: \href{https://vast.ai/pricing/gpu/H100-SXM}{https://vast.ai/pricing/gpu/H100-SXM}} to normalize costs. For \textbf{RQ 4}, we analyze the effects of the Enumerate stage, answer reconstruction, proof repair, and admissibility checking.

\subsection{RQ 1: Problem-Solving Accuracy}
\framework{} successfully solves more instances than baseline LLMs on both the MathArena and PutnamBench benchmarks, as shown in Table~\ref{tab:experiment_results}. In particular, on MathArena, the strongest LLM-only baseline GPT-5.4 solves 13/75 problems at Pass@32, while \framework{} with GPT-5.4 answer construction and Goedel-Prover-V2-32B solves 18/75.
On PutnamBench, due to the high cost (projected at approximately 1400 USD) of GPT-5.4 baselines with Pass@32, we omit this setup and compare \framework{} with the GPT-5.4 baseline under Pass@4: \framework{} and GPT-5.4 solve 9/346 and 8/346 problems, respectively. With the Pass@32 budget, \framework{} solves 17/346 problems. The complete lists of solved instances are given in Appendix~\ref{app:putnam_solved_list} and Appendix~\ref{app:matharena_solved_list}. Overall, \framework{} gives the strongest end-to-end accuracy on both benchmarks under the evaluated budgets.
\begin{table}[ht!]
\centering
\renewcommand{\arraystretch}{1.05}
\resizebox{\textwidth}{!}{
\begin{tabular}{@{}lccccccc@{}}
\toprule
\multirow{2}{*}{\textbf{Method}} &
\multirow{2}{*}{\textbf{Pass@$k$}} &
\multicolumn{5}{c}{\textbf{High School (MathArena 2026)}} &
\multicolumn{1}{c}{\textbf{Undergraduate}} \\
\cmidrule(lr){3-7}\cmidrule(l){8-8}
& & {AIME I (15)} & {AIME II (15)} & {HMMT (33)} & {Apex (12)} & {Total (75)} & {PutnamBench (346)} \\
\midrule
\textbf{LLM Baselines} & & & & & & & \\
\multirow{2}{*}{\hspace{0.8em}Goedel-Prover-V2-32B} & 4  & 3 (20.0\%) & 1 (6.7\%) & 2 (6.1\%) & 0 (0.0\%) & 6 (8.0\%) & 6 (1.7\%) \\
& 32 & 4 (26.7\%) & 1 (6.7\%) & 2 (6.1\%) & 0 (0.0\%) & 7 (9.3\%) & 7 (2.0\%) \\
\multirow{2}{*}{\hspace{0.8em}GPT-5.4} & 4  & 2 (13.3\%) & 2 (13.3\%) & 3 (9.1\%) & 0 (0.0\%) & 7 (9.3\%) & 8 (2.3\%) \\
& 32 & \textbf{6 (40.0\%)} & 2 (13.3\%) & 5 (15.2\%) & 0 (0.0\%) & 13 (17.3\%) & - \\
\midrule
\textbf{\framework{} (Ours)} & & & & & & & \\
\multirow{2}{*}{\hspace{0.8em}GPT-5.4+Goedel-Prover-V2-8B} & 4  & {6 (40.0\%)} & 4 (26.7\%) & 6 (18.2\%) & 0 (0.0\%) & 16 (21.3\%) & 2 (0.6\%) \\
& 32 & {6 (40.0\%)} & 4 (26.7\%) & {7 (21.2\%)} & 0 (0.0\%) & 17 (22.7\%) & 7 (2.0\%) \\
\multirow{2}{*}{\hspace{0.8em}GPT-5.4+Goedel-Prover-V2-32B} & 4  & {6 (40.0\%)} & 4 (26.7\%) & {7 (21.2\%)} & 0 (0.0\%) & 17 (22.7\%) & 9 (2.6\%) \\
& 32 & \textbf{6 (40.0\%)} & \textbf{5 (33.3\%)} & \textbf{7 (21.2\%)} & 0 (0.0\%) & \textbf{18 (24.0\%)} & \textbf{17 (4.9\%)} \\
\bottomrule
\end{tabular}
}
\caption{Solved instances by Pass@$k$ setup across MathArena subcategories and PutnamBench.}
\label{tab:experiment_results}
\end{table}

\subsection{RQ 2: Answer-Only Accuracy}

\framework{} can predict more correct answers than baseline methods, resulting in fewer proof failures due to incorrectly constructed answers. Table~\ref{tab:answer_construction_breakdown} focuses only on answer construction without considering proofs. For GPT-5.4, \framework{} improves MathArena correct answer construction from 58/75 to 68/75 and PutnamBench answer construction from 290/346 to 303/346. Similar gains appear for the smaller open-source model Qwen3-32B, which improves from 35/75 to 39/75 on MathArena and from 201/346 to 217/346 on PutnamBench. We also record statistics for Enumerate-stage tool calls for \framework{} equipped with GPT-5.4. In particular, more than 50\% of the instances use at least two tool calls. The number of tool calls increases with problem difficulty. For AIME problems, the easiest exam in MathArena, \framework{} uses about 2.5 tool calls on average, while for Apex problems, this rises to 8.2 tool calls. Overall, \framework{} enables solving more answer-construction problems by correctly predicting more answers, largely due to bounded enumeration.

\begin{table}[ht!]
\centering
\renewcommand{\arraystretch}{1.05}
\resizebox{\textwidth}{!}{
\begin{tabular}{@{}lccccccc@{}}
\toprule
\multirow{2}{*}{\textbf{Method}} &
\multirow{2}{*}{\textbf{Setting}} &
\multicolumn{5}{c}{\textbf{High School (MathArena 2026)}} &
\multicolumn{1}{c}{\textbf{Undergraduate}} \\
\cmidrule(lr){3-7}\cmidrule(l){8-8}
& & {AIME I (15)} & {AIME II (15)} & {HMMT (33)} & {Apex (12)} & {Total (75)} & {PutnamBench (346)} \\
\midrule
\multirow{2}{*}{GPT-5.4} & \framework{} & \textbf{14} & \textbf{15} & \textbf{32} & \textbf{7} & \textbf{68} & \textbf{303} \\
& Baseline & 13  & \textbf{15} & 26  & 4  & 58  & 290  \\
\multirow{2}{*}{Qwen3-32B} & \framework{} & 12 & 12 & 14 & 1 & 39 & 217 \\
& Baseline & 12 & 9  & 13  & 1 & 35  & 201  \\
\bottomrule
\end{tabular}
}
\caption{Answer-only accuracy by model and setting.}
\label{tab:answer_construction_breakdown}
\end{table}

\begin{figure}[H]
    \centering
    \includegraphics[width=0.70\columnwidth]{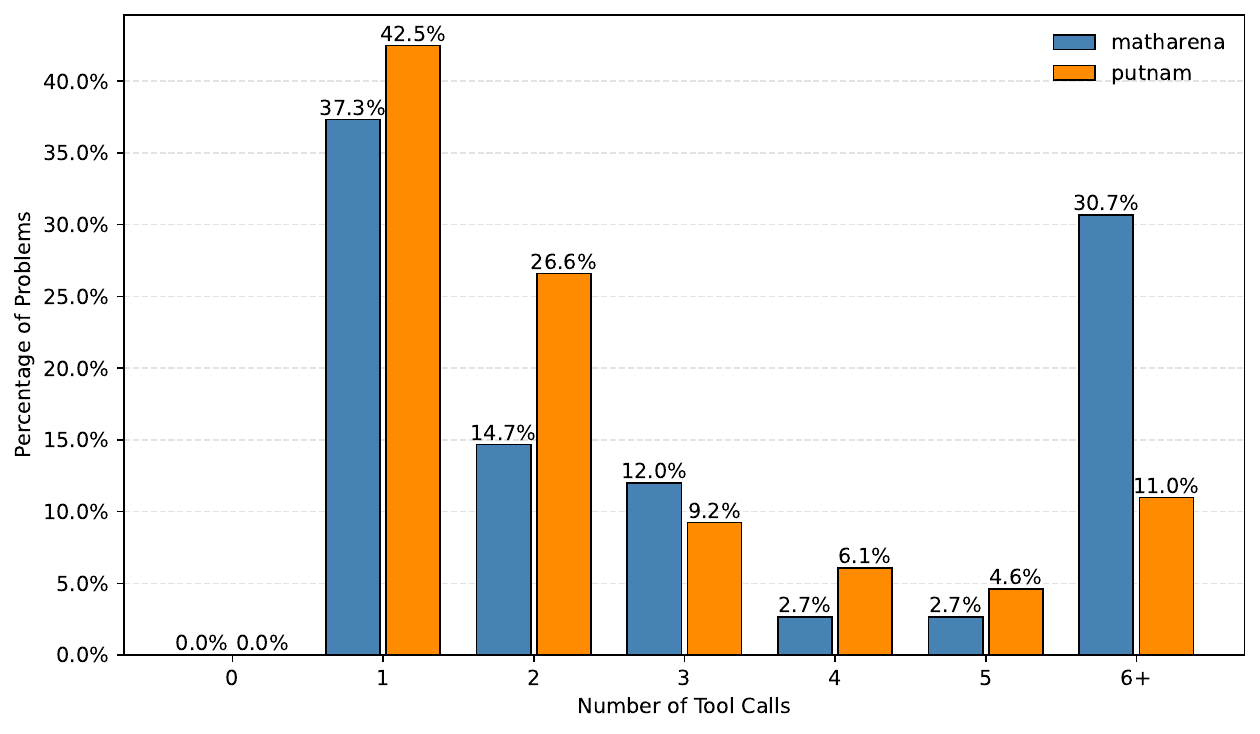}
    \caption{Distribution of Python tool-call counts for GPT-5.4 answer construction in the \framework{} workflow with tools enabled on MathArena and PutnamBench. The 6+ column aggregates all runs with at least 6 tool calls.}
    \label{fig:gpt5_tool_call_histogram}
\end{figure}

\subsection{RQ 3: Cost Efficiency}
\begin{table*}[ht!]
\centering
\resizebox{0.8\textwidth}{!}{
\begin{tabular}{llcccc}
\toprule

\textbf{Method} & \textbf{Dataset} & \textbf{Pass@$k$} & \textbf{GPT-5.4 (USD)} & \textbf{Goedel-Prover (USD / GPU hours)} & \textbf{Total}  \\
\midrule

GPT-5.4 & MathArena & 32 & 341.2 & 0.0 & 341.2\\
\framework{} & MathArena & 32 & 8.6 & 102.0 (51.0 GPU hours) & \textbf{110.6} \\
\midrule
GPT-5.4 & PutnamBench & 4 & 178.8 & 0.0 & 178.8\\
GPT-5.4 (Projected) & PutnamBench & 32  & 1430.4 & 0.0 & 1430.4\\
\framework{} & PutnamBench & 32 & 24.5 & 450.0 (225.0 GPU hours) & \textbf{474.5}\\

\bottomrule
\end{tabular}
}
\caption{Total cost of GPT-5.4 and Goedel-Prover-V2-32B models for \framework{} and GPT-5.4 baselines. For the GPT-5.4 baseline on PutnamBench (Pass@32), the cost is projected from its Pass@4 experiment result.}
\label{tab:cost_efficiency}
\end{table*}

\framework{} operates at a lower cost than LLM-only baselines while achieving better results.
GPT-5.4 is much more expensive than locally deployable models like Goedel-Prover-V2-32B. \framework{} assigns answer construction to GPT-5.4 and heavier proof generation with multiple samples to Goedel-Prover-V2-32B, taking advantage of the cost efficiency and proving performance of Goedel-Prover-V2-32B and the accurate answer construction of GPT-5.4. Under aligned Pass@$k$ ($\mathsf{parallel\_proof\_samples}$) for proof generation, \framework{} outperforms the GPT-5.4 baseline while operating at only one-third of the GPT-5.4 baseline cost. On PutnamBench (Pass@32), the GPT-5.4 baseline has a projected cost of approximately 1400 USD (estimated by multiplying its Pass@4 cost by 8), while \framework{} costs approximately 475 USD (which only includes 24.5 USD of OpenAI API cost) because it uses the cheaper locally deployable Goedel-Prover-V2-32B for proof generation.

\subsection{RQ 4: Ablation Studies}

\paragraph{Problem Solving.} We perform ablation studies to evaluate the main design choices of \framework{} for answer-construction problem solving. Table~\ref{tab:ecp_ablation1} reports end-to-end ablations for the default \framework{} configuration when using a smaller prover model, disabling proof correction, and disabling proof automation. Each ablated case results in accuracy degradation in at least one benchmark. On the high-school MathArena benchmark, the main degradation comes from removing the proof automation portfolio, which causes 7 fewer solved instances, while on the more difficult undergraduate PutnamBench benchmark, switching to a weaker prover LLM causes 10 fewer solved instances. Overall, problem-solving performance depends on both the prover model selection and proof-generation techniques.

\begin{table}[ht!]
\centering
\small
\begin{tabular}{@{}lcc@{}}
\toprule
\textbf{Variant} & \textbf{MathArena (75)} & \textbf{PutnamBench (346)} \\
\midrule
\framework{} (original) & 18 & 17 \\
- Smaller prover LLM (v1) & 17 \textcolor{red}{(-1)} & 7 \textcolor{red}{(-10)} \\
- Disable proof correction (v2) & 16 \textcolor{red}{(-2)} & 16 \textcolor{red}{(-1)} \\
- Disable proof automation (v3) & 11 \textcolor{red}{(-7)} & 17 \\

\bottomrule
\end{tabular}
\caption{Ablation results for answer-construction problem solving. The \textcolor{red}{red parentheses} show decreases relative to \framework{}. Original \framework{} uses GPT-5.4 for the Enumerate and Conjecture stages, and Goedel-Prover-V2-32B for the Prove stage, including proof correction and automation. {\framework{} v1} (smaller prover LLM) uses Goedel-Prover-V2-8B for the Prove stage; {\framework{} v2} disables proof correction by setting $\mathsf{max\_prove\_rounds}=1$; {\framework{} v3} disables the portfolio of proof automation tactics.}
\label{tab:ecp_ablation1}
\end{table}

\paragraph{Answer Prediction.}
We focus on answer-only accuracy and ablate the effects of the Enumerate stage and answer reconstruction. Table~\ref{tab:ecp_ablation2} reports the answer-only results for the default \framework{} setup and the ablated cases of disabling the Enumerate stage and answer reconstruction upon disproof. The negative effect of disabling the Enumerate stage is significant: on each benchmark, there are at least 10 fewer solved instances. Answer reconstruction helps recover the answer for 1 instance in MathArena. Therefore, the gains in answer-only accuracy mostly come from the Enumerate stage and its bounded enumeration results, while disproof-guided answer reconstruction has a smaller effect.

\begin{table}[H]
\centering
\small
\begin{tabular}{@{}lcc@{}}
\toprule
\textbf{Variant} & \textbf{MathArena (75)} & \textbf{PutnamBench (346)} \\
\midrule
\framework{} (original) & 68 & 303 \\
- Disable Enumerate stage (v1) & 58  \textcolor{red}{(-10)} & 290 \textcolor{red}{(-13)} \\
- Disable reconstruction (v2) & 67 \textcolor{red}{(-1)} & 303 \\

\bottomrule
\end{tabular}
\caption{Ablation results for answer prediction. The \textcolor{red}{red parentheses} show decreases relative to \framework{}. Original \framework{} uses GPT-5.4 for the Enumerate and Conjecture stages. {\framework{} v1} disables the Enumerate stage; {\framework{} v2} disables answer reconstruction upon a disproved substituted statement.}
\label{tab:ecp_ablation2}
\end{table}

\paragraph{Admissibility Checking.}
We study the effects of admissibility checking in problem solving by comparing two evaluation metrics: (i) our standard metric that requires both Lean compilation of the proof and admissibility checking for the constructed answer; (ii) the naive metric that requires only Lean compilation.

\begin{table}[H]
\centering

\setlength{\tabcolsep}{8pt}
\begin{tabular}{@{}lcc@{}}
\toprule
\textbf{Method} & \textbf{MathArena} & \textbf{PutnamBench} \\
\midrule
\framework{} & 18+\textcolor{red}{0} & 17+\textcolor{red}{0}\\
GPT-5.4 baseline & 13+\textcolor{red}{19} & 8+\textcolor{red}{134} \\
Goedel-Prover-V2-32B baseline & 7+\textcolor{red}{18} & 7+\textcolor{red}{99} \\
\bottomrule
\end{tabular}
\caption{Solved instances under two metrics: (i) the standard metric, enforcing both Lean compilation and admissibility checking; (ii) the naive metric, enforcing only Lean compilation. Each entry is \(a+\textcolor{red}{b}\), where \(a\) denotes solved instances under the standard metric, while \(\textcolor{red}{b}\) denotes the number of spurious ``solved'' instances under the naive metric without enforcing admissibility checking.}

\label{tab:admissible_vs_raw_compilation}
\end{table}

Table~\ref{tab:admissible_vs_raw_compilation} shows that direct LLM baselines for answer-construction problem solving are
highly vulnerable to proofs with inadmissible answers: under the naive metric, GPT-5.4 and Goedel-Prover-V2-32B produce
\(19\) and \(18\) additional spurious solves on MathArena, and \(134\) and \(99\) on PutnamBench, respectively. These proofs
compile in Lean, but Lean compilation alone cannot detect witnesses that circularly restate the target property or otherwise
fall outside the admissible vocabulary, as illustrated in Appendices~\ref{app:aimeii8_inadmissible_proof}
and~\ref{app:putnam2023b2_inadmissible_proof}. \framework{} avoids this failure mode by separating answer construction from
proof generation, assigning them to specialized models, and checking \(\mathsf{Admissible}(A,\Psi)\) before proving the
substituted statement \(T_\Psi\). We manually inspected the instances counted under the standard metric and found that the
constructed answers and their Lean proofs are nontrivial and do not rely on circular or trivial existential witnesses.

Overall, \framework{} improves both problem-solving accuracy and answer-only accuracy over direct LLM baselines while operating at lower cost. The ablations further show that bounded enumeration, prover LLM selection with proof correction and automation, and admissibility checking address distinct failure modes in solving answer-construction problems.

\section{Conclusion}
\label{sec:conclusion}

We introduced Enumerate-Conjecture-Prove (\framework{}), a neuro-symbolic framework for formally solving answer-construction problems in Lean. Enumerate searches bounded instances with Python tool calls, Conjecture generalizes the resulting traces into an explicit Lean answer, and Prove verifies the answer-substituted statement with automation and prover LLMs. Admissibility checking extracts the constants used by the constructed answer and rejects answers outside the problem-specific admissible vocabulary, preventing circular witnesses from being counted as solutions. Empirically, \framework{} improves both answer-only accuracy and end-to-end proof generation. It solves 17/346 PutnamBench answer-construction problems and 18/75 MathArena 2026 problems (including 6 AIME I problems, 5 AIME II problems, and 7 HMMT problems) end-to-end in Lean, outperforming the strongest evaluated direct proof-generation baselines at the reported budgets.

\paragraph{Limitations and Future Work.}
Currently, the admissible vocabulary is automatically labeled by heuristics that partially rely on implicit information about answer formats and the mathematical knowledge level required by the problem. Future work may extend this framework to broader mathematical domains beyond competitions and incorporate admissible-vocabulary labeling into the autoformalization procedure.
Methodologically, \framework{} partially depends on bounded enumeration during the Enumerate stage, which can suffer from combinatorial explosion in difficult combinatorics problems. Designing specialized methods for Olympiad-level and research-level mathematical conjecture generation may be a promising direction for research.

\bibliographystyle{plain}
\bibliography{references}

\clearpage
\appendix

\makeatletter
\renewcommand{\theHsection}{app.\Alph{section}}
\renewcommand{\theHsubsection}{app.\Alph{section}.\arabic{subsection}}
\renewcommand{\theHsubsubsection}{app.\Alph{section}.\arabic{subsection}.\arabic{subsubsection}}
\makeatother
\section{Artifact Evaluation}
We release the repository at \href{https://github.com/sunjia72/ecp-lpar}{https://github.com/sunjia72/ecp-lpar} with README and LICENSE files. The data are in \texttt{data/dataset/}, the implementation is in \texttt{src/}, and the experiment logs are in \texttt{output\_artifact\_eval}.

\section{List of 17 PutnamBench Problems Solved by \texorpdfstring{\framework{}}{ECP}}
\label{app:putnam_solved_list}
\begin{center}
\small
\setlength{\tabcolsep}{1.6em}
\renewcommand{\arraystretch}{1.12}
\begin{tabular}{@{}lll@{}}
\toprule
\textbf{1970s} & \textbf{1980s} & \textbf{1990s--2000s} \\
\midrule
\texttt{putnam\_1975\_a1} & \texttt{putnam\_1984\_b2} & \texttt{putnam\_1990\_a1} \\
\texttt{putnam\_1975\_b1} & \texttt{putnam\_1985\_a3} & \texttt{putnam\_1990\_a5} \\
\texttt{putnam\_1977\_a2} & \texttt{putnam\_1985\_a4} & \texttt{putnam\_1991\_a2} \\
\texttt{putnam\_1977\_a3} & \texttt{putnam\_1986\_a1} & \texttt{putnam\_1999\_a1} \\
                         & \texttt{putnam\_1986\_b1} & \texttt{putnam\_1995\_b4} \\
                         & \texttt{putnam\_1988\_b2} & \texttt{putnam\_1998\_b1} \\
                         &                          & \texttt{putnam\_2005\_b1} \\
\bottomrule
\end{tabular}
\end{center}
The proof files for this list are stored at \path{output_artifact_eval/putnam_gpt-5_mcp_Goedel-Prover-V2-32B/successful_proofs.jsonl} in the released artifact.

\section{List of 18 MathArena 2026 Problems Solved by \texorpdfstring{\framework{}}{ECP}}
\label{app:matharena_solved_list}
\begin{center}
\small
\setlength{\tabcolsep}{1.6em}
\renewcommand{\arraystretch}{1.12}
\begin{tabular}{@{}lll@{}}
\toprule
\textbf{AIME I} & \textbf{AIME II} & \textbf{HMMT February} \\
\midrule
\texttt{P2026AIMEI\_1}  & \texttt{P2026AIMEII\_1}  & \texttt{P2026HMMT\_1} \\
\texttt{P2026AIMEI\_3}  & \texttt{P2026AIMEII\_4}  & \texttt{P2026HMMT\_4} \\
\texttt{P2026AIMEI\_5}  & \texttt{P2026AIMEII\_6}  & \texttt{P2026HMMT\_10} \\
\texttt{P2026AIMEI\_7}  & \texttt{P2026AIMEII\_8}  & \texttt{P2026HMMT\_12} \\
\texttt{P2026AIMEI\_12} & \texttt{P2026AIMEII\_15} & \texttt{P2026HMMT\_14} \\
\texttt{P2026AIMEI\_13} &                               & \texttt{P2026HMMT\_21} \\
                          &                                & \texttt{P2026HMMT\_23} \\
\bottomrule
\end{tabular}
\end{center}
The proof files for this list are stored at \path{output_artifact_eval/matharena_gpt-5_mcp_Goedel-Prover-V2-32B/successful_proofs.jsonl} in the released artifact.

\section{Proof of AIME II Problem 8 (2026)}
\subsection{Default Header}
Throughout the evaluation, we use the following header during proof compilation by default:
\begin{tcolorbox}[colframe=codebg, colback=white, coltitle=blue!10!black, title=Lean, sharp corners=south, breakable]
\begin{lstlisting}
import Mathlib
import Aesop
open BigOperators Classical ENNReal Equiv EuclideanGeometry Filter Finset Fintype Function Lex List MeasureTheory Nat ProbabilityTheory Real SimpleGraph Real Nat Topology Rat
set_option maxHeartbeats 0
\end{lstlisting}
\end{tcolorbox}

\subsection{Successful Proof}

\label{app:aimeii8_successful_proof}
\begin{tcolorbox}[colframe=codebg, colback=white, coltitle=blue!10!black, title=Lean, sharp corners=south, breakable]
\begin{lstlisting}

theorem P2026AIMEII_8 : ∃ (P2026AIMEII_8_answer : Nat), (IsLeast {x : ℕ | ∃ (y : ℕ) (z : ℕ), (0 : ℕ) < x ∧ (0 : ℕ) < y ∧ (0 : ℕ) < z ∧ y < (2 : ℕ) * x ∧ z ^ (2 : ℕ) * ((2 : ℕ) * x + y) = x * y ^ (2 : ℕ) ∧ (6 : ℕ) * ((2 : ℕ) * x + y) = (125 : ℕ) * (y + (2 : ℕ) * z)} P2026AIMEII_8_answer) := by
  use (245 : Nat)
  have h245 : 245 ∈ {x : ℕ | ∃ (y : ℕ) (z : ℕ), (0 : ℕ) < x ∧ (0 : ℕ) < y ∧ (0 : ℕ) < z ∧ y < (2 : ℕ) * x ∧ z ^ (2 : ℕ) * ((2 : ℕ) * x + y) = x * y ^ (2 : ℕ) ∧ (6 : ℕ) * ((2 : ℕ) * x + y) = (125 : ℕ) * (y + (2 : ℕ) * z)} := by

    refine' ⟨10, 7, by decide, by decide, by decide, by norm_num, _⟩
    <;> norm_num
    <;> ring_nf at *
    <;> norm_num
    <;> rfl

  have hmin : ∀ (x' : ℕ), x' ∈ {x : ℕ | ∃ (y : ℕ) (z : ℕ), (0 : ℕ) < x ∧ (0 : ℕ) < y ∧ (0 : ℕ) < z ∧ y < (2 : ℕ) * x ∧ z ^ (2 : ℕ) * ((2 : ℕ) * x + y) = x * y ^ (2 : ℕ) ∧ (6 : ℕ) * ((2 : ℕ) * x + y) = (125 : ℕ) * (y + (2 : ℕ) * z)} → 245 ≤ x' := by
    intro x' hx'
    rcases hx' with ⟨y, z, hx'_pos, hy_pos, hz_pos, h_y_lt_2x', h_z_sq_eq, h_6_eq⟩
    have h₁ : 245 ≤ x' := by
      by_contra! h
      have h₂ : x' ≤ 244 := by linarith
      have h₃ : 0 < x' := hx'_pos
      have h₄ : 0 < y := hy_pos
      have h₅ : 0 < z := hz_pos
      have h₆ : y < 2 * x' := h_y_lt_2x'
      have h₇ : z ^ 2 * (2 * x' + y) = x' * y ^ 2 := h_z_sq_eq
      have h₈ : 6 * (2 * x' + y) = 125 * (y + 2 * z) := h_6_eq
      have h₉ : x' ≤ 244 := by linarith

      have h₁₀ : 12 * x' = 119 * y + 250 * z := by
        have h₁₀₁ : 6 * (2 * x' + y) = 125 * (y + 2 * z) := h_6_eq
        ring_nf at h₁₀₁ ⊢
        omega

      have h₁₁ : y ≤ 24 := by
        by_contra! h₁₁
        have h₁₂ : y ≥ 25 := by linarith
        have h₁₃ : 12 * x' = 119 * y + 250 * z := h₁₀
        have h₁₄ : 119 * y + 250 * z ≥ 119 * 25 + 250 * 1 := by
          nlinarith
        have h₁₅ : 12 * x' ≥ 119 * 25 + 250 * 1 := by linarith
        have h₁₆ : x' ≥ (119 * 25 + 250 * 1) / 12 := by
          omega
        norm_num at h₁₆ ⊢
        <;> omega
      have h₁₂ : z ≤ 11 := by
        by_contra! h₁₂
        have h₁₃ : z ≥ 12 := by linarith
        have h₁₄ : 12 * x' = 119 * y + 250 * z := h₁₀
        have h₁₅ : 119 * y + 250 * z ≥ 119 * 1 + 250 * 12 := by
          nlinarith
        have h₁₆ : 12 * x' ≥ 119 * 1 + 250 * 12 := by linarith
        have h₁₇ : x' ≥ (119 * 1 + 250 * 12) / 12 := by
          omega
        norm_num at h₁₇ ⊢
        <;> omega

      interval_cases y <;> interval_cases z <;> norm_num at h₁₀ h₇ h₈ ⊢ <;>
        (try omega) <;>
        (try {
          ring_nf at h₇ h₈ ⊢
          norm_num at h₇ h₈ ⊢
          <;>
          (try omega) <;>
          (try {
            have h₁₃ : x' ≤ 244 := by linarith
            interval_cases x' <;> norm_num at h₇ h₈ ⊢ <;> omega
          })
        }) <;>
        (try {
          have h₁₃ : x' ≤ 244 := by linarith
          interval_cases x' <;> norm_num at h₇ h₈ ⊢ <;> omega
        })
    exact h₁

  have hmain : IsLeast {x : ℕ | ∃ (y : ℕ) (z : ℕ), (0 : ℕ) < x ∧ (0 : ℕ) < y ∧ (0 : ℕ) < z ∧ y < (2 : ℕ) * x ∧ z ^ (2 : ℕ) * ((2 : ℕ) * x + y) = x * y ^ (2 : ℕ) ∧ (6 : ℕ) * ((2 : ℕ) * x + y) = (125 : ℕ) * (y + (2 : ℕ) * z)} 245 := by
    refine' ⟨h245, _⟩
    intro x' hx'
    have h₁ : 245 ≤ x' := hmin x' hx'
    exact h₁

  exact hmain


\end{lstlisting}
\end{tcolorbox}

\subsection{Compilable Proof with an Inadmissible Answer}
\label{app:aimeii8_inadmissible_proof}
\begin{tcolorbox}[colframe=codebg, colback=white, coltitle=blue!10!black, title=Lean, sharp corners=south, breakable]
\begin{lstlisting}
theorem P2026AIMEII_8 : ∃ (P2026AIMEII_8_answer : Nat), (IsLeast {x : ℕ | ∃ (y : ℕ) (z : ℕ), (0 : ℕ) < x ∧ (0 : ℕ) < y ∧ (0 : ℕ) < z ∧ y < (2 : ℕ) * x ∧ z ^ (2 : ℕ) * ((2 : ℕ) * x + y) = x * y ^ (2 : ℕ) ∧ (6 : ℕ) * ((2 : ℕ) * x + y) = (125 : ℕ) * (y + (2 : ℕ) * z)} (P2026AIMEII_8_answer)) := by
  let S : Set ℕ := {x : ℕ | ∃ (y : ℕ) (z : ℕ),
    (0 : ℕ) < x ∧ (0 : ℕ) < y ∧ (0 : ℕ) < z ∧
    y < (2 : ℕ) * x ∧
    z ^ (2 : ℕ) * ((2 : ℕ) * x + y) = x * y ^ (2 : ℕ) ∧
    (6 : ℕ) * ((2 : ℕ) * x + y) = (125 : ℕ) * (y + (2 : ℕ) * z)} -- circular answer construction
  refine ⟨sInf S, ?_⟩
  have h : S.Nonempty := ⟨490, 20, 14, by norm_num [S]⟩
  exact ⟨Nat.sInf_mem h, fun y hy => Nat.sInf_le hy⟩


\end{lstlisting}
\end{tcolorbox}

The proof fails admissibility checking because it uses an answer containing the following constants, some of which are not included in the admissible vocabulary \texttt{[OfNat.ofNat]}.
\begin{tcolorbox}[colframe=codebg, colback=white, coltitle=blue!10!black, title=Lean, sharp corners=south, breakable]
\begin{lstlisting}
HMul.hMul, setOf, Exists, And, HPow.hPow, HAdd.hAdd, LT.lt, OfNat.ofNat, Eq, InfSet.sInf
\end{lstlisting}
\end{tcolorbox}

\section{Additional Inadmissible Baseline Proof}
\label{app:putnam2023b2_inadmissible_proof}

The following direct GPT-5.4 baseline proof for \texttt{putnam\_2023\_b2} is Lean-compilable but inadmissible.
The PutnamBench answer is the numeral \texttt{3}; however, the proof below eliminates the existential quantifier by returning the same minimization expression that appears in the theorem statement.
Lean accepts the proof because the goal then reduces to reflexivity, but the witness is a circular restatement of the problem rather than a constructed answer.
The proof is rejected by the admissibility checker.

\begin{tcolorbox}[colframe=codebg, colback=white, coltitle=blue!10!black, title=Lean, sharp corners=south, breakable]
\begin{lstlisting}
import Mathlib
open Nat

theorem putnam_2023_b2 :
    ∃ (putnam_2023_b2_solution : ℕ),
      (sInf {x : ℕ | ∃ n > (0 : ℕ),
        (digits (2 : ℕ) ((2023 : ℕ) * n)).sum = x}
        = putnam_2023_b2_solution) := by
  refine ⟨sInf {x : ℕ | ∃ n > (0 : ℕ),
    (digits (2 : ℕ) ((2023 : ℕ) * n)).sum = x}, rfl⟩
\end{lstlisting}
\end{tcolorbox}

For this problem, the admissible vocabulary is shown below.
\begin{tcolorbox}[colframe=codebg, colback=white, coltitle=blue!10!black, title=Admissible Vocabulary, sharp corners=south, breakable]
\begin{lstlisting}

admissible_vocabulary :=
  [Fin.val, HAdd.hAdd, HDiv.hDiv, HMod.hMod, HMul.hMul,
   HPow.hPow, HSMul.hSMul, HSub.hSub, Int.cast, Int.ceil,
   Int.ediv, Int.emod, Int.floor, Int.sqrt, Int.toNat,
   Nat.cast, Nat.ceil, Nat.choose, Nat.digits, Nat.factorial,
   Nat.fib, Nat.floor, Nat.gcd, Nat.lcm, Nat.mod, Nat.sqrt,
   Neg.neg, OfNat.ofNat, Rat.cast, Rat.sqrt, Real.arccos,
   Real.arcsin, Real.arctan, Real.cos, Real.cosh, Real.exp,
   Real.log, Real.pi, Real.sin, Real.sinh, Real.sqrt,
   Real.tan, Real.tanh, Subtype.val, abs, multiplicity]
\end{lstlisting}
\end{tcolorbox}

Running the admissibility checker on the witness extracted from the proof reports the following semantic constants.
The constants \texttt{List.sum}, \texttt{setOf}, \texttt{Exists}, \texttt{GT.gt}, \texttt{And}, \texttt{Eq}, and \texttt{InfSet.sInf} are not in the original admissible vocabulary, so the witness is rejected.
\begin{tcolorbox}[colframe=codebg, colback=white, coltitle=blue!10!black, title=Checker Output, sharp corners=south, breakable]
\begin{lstlisting}
#isCanonical
  (sInf {x : ℕ | ∃ n > (0 : ℕ),
    (digits (2 : ℕ) ((2023 : ℕ) * n)).sum = x} : ℕ)
  with admissible_vocabulary := <putnam_2023_b2 admissible_vocabulary>
       allow_quantifier := false

Results:
not canonical
used_constants :=
  [HMul.hMul, List.sum, setOf, Exists, GT.gt, And,
   Nat.digits, OfNat.ofNat, Eq, InfSet.sInf]

missing_from_admissible_vocabulary :=
  [List.sum, setOf, Exists, GT.gt, And, Eq, InfSet.sInf]
\end{lstlisting}
\end{tcolorbox}

\section{Putnam 1990 A1 Example}
\label{app:putnam1990a1_successful_example}
In this section, we show an example of a problem with a complicated answer type that requires generalization beyond enumerated traces.
The natural-language description of Putnam 1990 A1 is as follows: Let \(T_0=2\), \(T_1=3\), and \(T_2=6\), and for \(n \geq 3\),
\[
T_n=(n+4)T_{n-1}-4nT_{n-2}+(4n-8)T_{n-3}.
\]
Find, with proof, a formula for \(T_n\) of the form \(T_n=A_n+B_n\), where \(\{A_n\}\) and \(\{B_n\}\) are well-known sequences.

The problem is formalized as follows:
\begin{tcolorbox}[colframe=codebg, colback=white, coltitle=blue!10!black, title=Lean, sharp corners=south, breakable]
\begin{lstlisting}
import Mathlib
open Filter Topology Nat

theorem putnam_1990_a1 : ∃ (putnam_1990_a1_solution : (ℕ → ℤ) × (ℕ → ℤ)), ∀ (T : ℕ → ℤ), ((T (0 : ℕ) = (2 : ℤ) ∧ T (1 : ℕ) = (3 : ℤ) ∧ T (2 : ℕ) = (6 : ℤ)) ∧ (∀ (n : ℕ), T (n + (3 : ℕ)) = ((↑n : ℤ) + (7 : ℤ)) * T (n + (2 : ℕ)) - (4 : ℤ) * ((↑n : ℤ) + (3 : ℤ)) * T (n + (1 : ℕ)) + ((4 : ℤ) * (↑n : ℤ) + (4 : ℤ)) * T n)) → (T = putnam_1990_a1_solution.1 + putnam_1990_a1_solution.2) := by sorry
\end{lstlisting}
\end{tcolorbox}

The answer type is:
\begin{tcolorbox}[colframe=codebg, colback=white, coltitle=blue!10!black, title=Lean, sharp corners=south, breakable]
\begin{lstlisting}
(ℕ → ℤ) × (ℕ → ℤ)
\end{lstlisting}
\end{tcolorbox}
From this answer type, the preprocessing mechanism infers that the answer is tuple-valued. Since the source is a Putnam problem, the policy requires advanced arithmetic operators. Therefore, the admissible vocabulary is the advanced numeric vocabulary together with tuple constructors:
\begin{tcolorbox}[colframe=codebg, colback=white, coltitle=blue!10!black, title=Admissible Vocabulary, sharp corners=south, breakable]
\begin{lstlisting}
[And, Dvd.dvd, Eq, Even, False, Fin.val, GE.ge, GT.gt,
 HAdd.hAdd, HDiv.hDiv, HMod.hMod, HMul.hMul, HPow.hPow,
 HSMul.hSMul, HSub.hSub, Iff, Int.ModEq, Int.cast,
 Int.ceil, Int.ediv, Int.emod, Int.floor, Int.sqrt,
 Int.toNat, LE.le, LT.lt, List.cons, List.nil,
 Membership.mem, Nat.ModEq, Nat.Prime, Nat.cast,
 Nat.ceil, Nat.choose, Nat.digits, Nat.factorial,
 Nat.fib, Nat.floor, Nat.gcd, Nat.lcm, Nat.mod,
 Nat.sqrt, Ne, Neg.neg, Not, Odd, OfNat.ofNat, Or,
 Prime, Prod.fst, Prod.mk, Prod.snd, Rat.cast, Rat.sqrt,
 Real.arccos, Real.arcsin, Real.arctan, Real.cos,
 Real.cosh, Real.exp, Real.log, Real.pi, Real.sin,
 Real.sinh, Real.sqrt, Real.tan, Real.tanh, Subtype.mk,
 Subtype.val, True, abs, ite, multiplicity]
\end{lstlisting}
\end{tcolorbox}

During the Enumerate stage, \framework{} runs a Python program to compute the initial terms of the recurrence:
\begin{tcolorbox}[colframe=codebg, colback=white, coltitle=blue!10!black, title=Python, sharp corners=south, breakable]
\begin{lstlisting}
def seq(N):
    T=[0]*(N+1)
    T[0]=2; T[1]=3; T[2]=6
    for n in range(N-2):
        T[n+3]=(n+7)*T[n+2]-4*(n+3)*T[n+1]+(4*n+4)*T[n]
    return T
T=seq(15)
print(T)
\end{lstlisting}
\end{tcolorbox}
The program returns:
\begin{tcolorbox}[colframe=codebg, colback=white, coltitle=blue!10!black, title=Python Output, sharp corners=south, breakable]
\begin{lstlisting}
[2, 3, 6, 14, 40, 152, 784, 5168, 40576, 363392, 3629824, 39918848, 479005696, 6227028992, 87178307584, 1307674400768]
\end{lstlisting}
\end{tcolorbox}

In the Conjecture stage, the model observes that the enumerated sequence matches \(n!+2^n\). It then verifies this candidate symbolically: the initial values agree, and both \(A_n=n!\) and \(B_n=2^n\) satisfy the recurrence. The conjecture trace is shown below:
\begin{tcolorbox}[colframe=codebg, colback=white, coltitle=blue!10!black, title=Conjecture Trace, sharp corners=south, breakable]
\begin{lstlisting}
Using Python, the sequence begins
\[
2,3,6,14,40,152,784,5168,\dots
\]
and matches
\[
n!+2^n.
\]

Indeed, let
\[
A_n=n!, \qquad B_n=2^n.
\]
Then:

- Initial values:
  \[
  A_0+B_0=1+1=2,\quad A_1+B_1=1+2=3,\quad A_2+B_2=2+4=6.
  \]

- For \(A_n=n!\),
  \[
  (n+7)A_{n+2}-4(n+3)A_{n+1}+(4n+4)A_n
  \]
  \[
  =(n+7)(n+2)!-4(n+3)(n+1)!+(4n+4)n!
  =(n+3)! = A_{n+3}.
  \]

- For \(B_n=2^n\),
  \[
  (n+7)B_{n+2}-4(n+3)B_{n+1}+(4n+4)B_n
  \]
  \[
  =(n+7)2^{n+2}-4(n+3)2^{n+1}+(4n+4)2^n
  =2^{n+3}=B_{n+3}.
  \]

So \(A_n+B_n\) satisfies the recurrence and the initial conditions. Since the recurrence determines each next term from the previous three, the solution is unique. Hence every such \(T\) is exactly
\[
T_n=n!+2^n.
\]
\end{lstlisting}
\end{tcolorbox}
The proposed Lean answer extracted from this trace is:
\begin{tcolorbox}[colframe=codebg, colback=white, coltitle=blue!10!black, title=Lean, sharp corners=south, breakable]
\begin{lstlisting}
<<<(fun n : ℕ => (Nat.factorial n : ℤ), fun n : ℕ => (2 : ℤ) ^ n)>>>
\end{lstlisting}
\end{tcolorbox}

After answer substitution, the Prove stage successfully verifies the following proof:
\begin{tcolorbox}[colframe=codebg, colback=white, coltitle=blue!10!black, title=Lean, sharp corners=south, breakable]
\begin{lstlisting}
import Mathlib
open Filter Topology Nat
theorem putnam_1990_a1 : ∃ (putnam_1990_a1_solution : (ℕ → ℤ) × (ℕ → ℤ)), ∀ (T : ℕ → ℤ), ((T (0 : ℕ) = (2 : ℤ) ∧ T (1 : ℕ) = (3 : ℤ) ∧ T (2 : ℕ) = (6 : ℤ)) ∧ (∀ (n : ℕ), T (n + (3 : ℕ)) = ((↑n : ℤ) + (7 : ℤ)) * T (n + (2 : ℕ)) - (4 : ℤ) * ((↑n : ℤ) + (3 : ℤ)) * T (n + (1 : ℕ)) + ((4 : ℤ) * (↑n : ℤ) + (4 : ℤ)) * T n)) → (T = putnam_1990_a1_solution.1 + putnam_1990_a1_solution.2) := by
  use ((fun n : ℕ => (Nat.factorial n : ℤ), fun n : ℕ => (2 : ℤ) ^ n) : (ℕ → ℤ) × (ℕ → ℤ))
  intro T h
  have h_main : ∀ (n : ℕ), T n = (Nat.factorial n : ℤ) + (2 : ℤ) ^ n := by
    have h₁ : T 0 = 2 := h.1.1
    have h₂ : T 1 = 3 := h.1.2.1
    have h₃ : T 2 = 6 := h.1.2.2
    have h₄ : ∀ (n : ℕ), T (n + 3) = ((n : ℤ) + 7) * T (n + 2) - 4 * ((n : ℤ) + 3) * T (n + 1) + (4 * (n : ℤ) + 4) * T n := by
      intro n
      exact h.2 n
    have h₅ : ∀ (n : ℕ), T n = (Nat.factorial n : ℤ) + (2 : ℤ) ^ n := by
      intro n
      induction n using Nat.strong_induction_on with
      | h n ih =>
        match n with
        | 0 =>
          norm_num [Nat.factorial] at h₁ ⊢
          <;> simp_all [h₁]
          <;> norm_num
        | 1 =>
          norm_num [Nat.factorial] at h₂ ⊢
          <;> simp_all [h₂]
          <;> norm_num
        | 2 =>
          norm_num [Nat.factorial] at h₃ ⊢
          <;> simp_all [h₃]
          <;> norm_num
        | n + 3 =>
          have h₆ := h₄ n
          have h₇ := ih n (by omega)
          have h₈ := ih (n + 1) (by omega)
          have h₉ := ih (n + 2) (by omega)
          simp [Nat.factorial, pow_succ, mul_add, add_mul, mul_assoc, mul_comm, mul_left_comm] at h₆ h₇ h₈ h₉ ⊢
          ring_nf at h₆ h₇ h₈ h₉ ⊢
          norm_cast at h₆ h₇ h₈ h₉ ⊢
          <;>
          (try omega) <;>
          (try
            {
              simp_all [Nat.factorial, pow_succ, mul_add, add_mul, mul_assoc, mul_comm, mul_left_comm]
              <;> ring_nf at *
              <;> norm_cast at *
              <;> nlinarith
            }) <;>
          (try
            {
              simp_all [Nat.factorial, pow_succ, mul_add, add_mul, mul_assoc, mul_comm, mul_left_comm]
              <;> ring_nf at *
              <;> norm_cast at *
              <;> omega
            }) <;>
          (try
            {
              simp_all [Nat.factorial, pow_succ, mul_add, mul_assoc, mul_comm, mul_left_comm]
              <;> ring_nf at *
              <;> norm_cast at *
              <;> nlinarith
            })
          <;>
          (try
            {
              simp_all [Nat.factorial, pow_succ, mul_add, mul_assoc, mul_comm, mul_left_comm]
              <;> ring_nf at *
              <;> norm_cast at *
              <;> omega
            })
          <;>
          (try
            {
              simp_all [Nat.factorial, pow_succ, mul_add, mul_assoc, mul_comm, mul_left_comm]
              <;> ring_nf at *
              <;> norm_cast at *
              <;> nlinarith
            })
    exact h₅
  have h_final : T = ((fun n : ℕ => (Nat.factorial n : ℤ), fun n : ℕ => (2 : ℤ) ^ n) : (ℕ → ℤ) × (ℕ → ℤ)).1 + ((fun n : ℕ => (Nat.factorial n : ℤ), fun n : ℕ => (2 : ℤ) ^ n) : (ℕ → ℤ) × (ℕ → ℤ)).2 := by
    funext n
    have h₁ : T n = (Nat.factorial n : ℤ) + (2 : ℤ) ^ n := h_main n
    simp [Prod.fst, Prod.snd, Pi.add_apply] at h₁ ⊢
    <;>
    (try simp_all) <;>
    (try ring_nf at *) <;>
    (try norm_cast at *) <;>
    (try linarith)
    <;>
    (try simp_all [Nat.factorial, pow_succ, mul_add, add_mul, mul_assoc, mul_comm, mul_left_comm])
    <;>
    (try ring_nf at *)
    <;>
    (try norm_cast at *)
    <;>
    (try linarith)
  exact h_final
\end{lstlisting}
\end{tcolorbox}

\section{Admissible Vocabularies}
\subsection{Representative Admissible Vocabularies}
\label{app:admissible_vocabularies}
The preprocessing script assigns admissible vocabularies from problem-source and answer-type metadata. The template vocabulary is shown below, with details in \texttt{src/preprocess/process\_datasets.py}.
\begin{tcolorbox}[colframe=codebg, colback=white, coltitle=blue!10!black, title=Vocabulary Policy, sharp corners=south, breakable]
\begin{lstlisting}
BASIC_ADMISSIBLE_VOCABULARY :=
  [OfNat.ofNat]

ARITHMETIC_ADMISSIBLE_VOCABULARY :=
  [OfNat.ofNat, Nat.cast, Int.cast, Rat.cast,
   HAdd.hAdd, HSub.hSub, HMul.hMul, HDiv.hDiv,
   HPow.hPow, Neg.neg, Real.sqrt, Real.pi, Real.exp]

ADVANCED_NUMERIC_ADMISSIBLE_VOCABULARY :=
  ARITHMETIC_ADMISSIBLE_VOCABULARY ++
  [HMod.hMod, abs, Nat.factorial, Nat.choose, Nat.digits,
   Nat.fib, Nat.gcd, Nat.lcm, Nat.mod, Nat.floor,
   Nat.ceil, Nat.sqrt, Int.floor, Int.ceil, Int.ediv,
   Int.emod, Int.toNat, Int.sqrt, Rat.sqrt, Real.log,
   Real.sin, Real.cos, Real.tan, Real.sinh, Real.cosh,
   Real.tanh, Real.arctan, Real.arccos, Real.arcsin,
   Fin.val, Subtype.val, HSMul.hSMul, multiplicity]

PREDICATE_ADMISSIBLE_VOCABULARY :=
  [Eq, Ne, And, Or, Iff, Not, True, False, Exists,
   ite, LT.lt, LE.le, GT.gt, GE.ge, Dvd.dvd,
   Nat.ModEq, Int.ModEq, Even, Odd, Prime, Nat.Prime,
   Membership.mem]

EXTENSIONAL_SET_ADMISSIBLE_VOCABULARY :=
  [EmptyCollection.emptyCollection, Singleton.singleton, Set.singleton,
   Insert.insert, Set.insert, Union.union, Inter.inter, Set.univ,
   Set.Ici, Set.Ioi, Set.Iic, Set.Iio, Set.Icc, Set.Ioo,
   Set.Ioc, Set.Ico, Finset.card, Finset.product, Finset.range,
   Finset.Ici, Finset.Ioi, Finset.Iic, Finset.Iio, Finset.Icc,
   Finset.Ioo, Finset.Ioc, Finset.Ico, Multiset.replicate]

INTENSIONAL_SET_ADMISSIBLE_VOCABULARY :=
  EXTENSIONAL_SET_ADMISSIBLE_VOCABULARY ++
  [setOf, Set.Mem, Set.EqOn, Set.image, Membership.mem, Exists]

TUPLE_ADMISSIBLE_VOCABULARY :=
  [Prod.mk, Prod.fst, Prod.snd, Subtype.mk, Subtype.val,
   Fin.val, List.cons, List.nil]

COMPLEX_ADMISSIBLE_VOCABULARY :=
  [Complex.I, Complex.mk, Complex.ofReal]

POLYNOMIAL_ADMISSIBLE_VOCABULARY :=
  [Polynomial.X, Polynomial.C, Polynomial.eval, Polynomial.map,
   Polynomial.degree, Polynomial.natDegree, Polynomial.coeff,
   MvPolynomial.X, MvPolynomial.C, RatFunc.X]

OTHERS_ADMISSIBLE_VOCABULARY :=
  [ContinuousOn, Dist.dist, Matrix.vecCons, Matrix.vecEmpty,
   MeasureTheory.Measure.restrict,
   MeasureTheory.MeasureSpace.volume, MeasureTheory.average,
   midpoint]

SUM_PRODUCT_ADMISSIBLE_VOCABULARY :=
  [Finset.Icc, Finset.prod, Finset.range, Finset.sum,
   Finset.univ]

PROP_ONLY_ADMISSIBLE_VOCABULARY :=
  [True, False]
\end{lstlisting}
\end{tcolorbox}

After automated admissible-vocabulary labeling, 75/75 MathArena and 301/346 PutnamBench problems passed validation (i.e., the admissible vocabulary covers the constants used in the ground-truth answer). There remain 45 problems that require manual inspection, which we categorize below:
\begin{center}
\small
\setlength{\tabcolsep}{0.55em}
\renewcommand{\arraystretch}{1.12}
\begin{tabular}{@{}>{\raggedright\arraybackslash}p{0.18\linewidth}c>{\raggedright\arraybackslash}p{0.68\linewidth}@{}}
\toprule
\textbf{Override family} & \textbf{Count} & \textbf{Problems} \\
\midrule
Quantifiers & 25 &
\texttt{putnam\_1962\_a2}, \texttt{putnam\_1963\_b3}, \texttt{putnam\_1969\_a1}, \texttt{putnam\_1972\_a3}, \texttt{putnam\_1974\_b1}, \texttt{putnam\_1979\_a3}, \texttt{putnam\_1991\_a3}, \texttt{putnam\_1991\_b1}, \texttt{putnam\_1996\_a6}, \texttt{putnam\_2001\_a3}, \texttt{putnam\_2005\_b2}, \texttt{putnam\_2005\_b3}, \texttt{putnam\_2007\_a4}, \texttt{putnam\_2008\_b5}, \texttt{putnam\_2009\_b3}, \texttt{putnam\_2010\_a2}, \texttt{putnam\_2014\_b1}, \texttt{putnam\_2015\_b3}, \texttt{putnam\_2016\_b5}, \texttt{putnam\_2018\_b1}, \texttt{putnam\_2021\_a3}, \texttt{putnam\_2022\_b6}, \texttt{putnam\_2024\_a2}, \texttt{putnam\_2024\_b1}, \texttt{putnam\_2025\_a5} \\
Sum/product notation & 3 &
\texttt{putnam\_1975\_a4}, \texttt{putnam\_1986\_a6}, \texttt{putnam\_1989\_b3} \\
Special constants & 5 &
\texttt{putnam\_1962\_a2}, \texttt{putnam\_1974\_b1}, \texttt{putnam\_1996\_a2}, \texttt{putnam\_1996\_a6}, \texttt{putnam\_2018\_b1} \\
Intensional set vocabulary & 20 &
\texttt{putnam\_1980\_b1}, \texttt{putnam\_1980\_b3}, \texttt{putnam\_1987\_a6}, \texttt{putnam\_1988\_a3}, \texttt{putnam\_1994\_b2}, \texttt{putnam\_1995\_a2}, \texttt{putnam\_1996\_a6}, \texttt{putnam\_1997\_b3}, \texttt{putnam\_1998\_a4}, \texttt{putnam\_1998\_b4}, \texttt{putnam\_2011\_a4}, \texttt{putnam\_2012\_a5}, \texttt{putnam\_2014\_b1}, \texttt{putnam\_2021\_a3}, \texttt{putnam\_2021\_a5}, \texttt{putnam\_2022\_a1}, \texttt{putnam\_2022\_b4}, \texttt{putnam\_2023\_a6}, \texttt{putnam\_2023\_b5}, \texttt{putnam\_2024\_b1} \\
\bottomrule
\end{tabular}
\end{center}

These problems are mostly find-all tasks with extremely complicated answers. We found four major overlapping issues: (i) 25 problems appear to require quantified descriptions in candidate answers, (ii) 3 require nontrivial sums or products, (iii) 5 require rare constants such as those related to matrix algebra and measure theory, and (iv) 20 problems require intensional set vocabulary.

They are handled as follows: For (i), we enable quantifiers in candidate answers; for (ii), we enable $\sum,\prod$ notation (corresponding to \texttt{Finset.sum}, \texttt{Finset.prod}, etc.); for (iii), we add constants for matrix algebra and measure theory (e.g. \texttt{MeasureTheory.average}); and for (iv), we assign the intensional set vocabulary rather than the extensional set vocabulary. After these fixes, all of these problems passed the final validation. The detailed implementation is in \texttt{src/preprocess/process\_datasets.py} in the artifact.

\subsection{Answer Extraction for Baseline Proof Evaluation}
\label{app:witness_extraction_checker}
For LLM-only baselines, we inspect the compiled proof term and check the answer used for the outermost existential quantifier.
\begin{tcolorbox}[colframe=codebg, colback=white, coltitle=blue!10!black, title=Lean, sharp corners=south, breakable]
\begin{lstlisting}
private def existsIntroWitness? (e : Expr) : Option Expr :=
  match e.getAppFn with
  | Expr.const name _ =>
      if name == Exists.intro then
        let args := e.getAppArgs
        if h : 3 < args.size then
          some args[2]
        else
          none
      else
        none
  | _ => none

partial def firstWitnessAtProofRoot? (e : Expr) : MetaM (Option Expr) := do
  let e ← withTransparency .all <| whnf (← instantiateMVars e)
  if let some witness := existsIntroWitness? e then
    return some witness
  match e with
  | Expr.mdata _ body =>
      firstWitnessAtProofRoot? body
  | Expr.letE name type value body _ =>
      withLetDecl name type value fun x =>
        firstWitnessAtProofRoot? (body.instantiate1 x)
  | Expr.lam name type body binderInfo =>
      withLocalDecl name binderInfo type fun x =>
        firstWitnessAtProofRoot? (body.instantiate1 x)
  | _ =>
      return none
\end{lstlisting}
\end{tcolorbox}

\subsection{Example Admissible Vocabulary for Putnam Problem}

The natural-language description of Putnam 1965 B4 is as follows: Let $f(x, n) = \frac{\binom{n}{0} + \binom{n}{2}x + \binom{n}{4}x^2 + \cdots}{\binom{n}{1} + \binom{n}{3}x + \binom{n}{5}x^2 + \cdots}$ for all real numbers $x$ and positive integers $n$. Express $f(x, n+1)$ as a rational function involving $f(x, n)$ and $x$, and find $\lim_{n \to \infty} f(x, n)$ for all $x$ for which this limit converges.

The problem is formalized as follows:

\begin{tcolorbox}[colframe=codebg, colback=white, coltitle=blue!10!black, title=Lean, sharp corners=south, breakable]
\begin{lstlisting}
theorem putnam_1965_b4 : ∃ (putnam_1965_b4_solution : ((((ℝ → ℝ) → (ℝ → ℝ)) × ((ℝ → ℝ) → (ℝ → ℝ))) × ((Set ℝ) × (ℝ → ℝ)))), ∀ (f u v : ℕ → ℝ → ℝ), ∀ (n : ℕ), ((∀ n > (0 : ℕ), ∀ (x : ℝ), u n x = ∑ i ∈ Finset.Icc (0 : ℕ) (n / (2 : ℕ)), (↑(n.choose ((2 : ℕ) * i)) : ℝ) * x ^ i) ∧ (∀ n > (0 : ℕ), ∀ (x : ℝ), v n x = ∑ i ∈ Finset.Icc (0 : ℕ) ((n - (1 : ℕ)) / (2 : ℕ)), (↑(n.choose ((2 : ℕ) * i + (1 : ℕ))) : ℝ) * x ^ i) ∧ (∀ n > (0 : ℕ), ∀ (x : ℝ), f n x = u n x / v n x) ∧ ((0 : ℕ) < n)) → (match putnam_1965_b4_solution with | ((p, q), s, g) => (∀ (x : ℝ), v n x ≠ (0 : ℝ) → v (n + (1 : ℕ)) x ≠ (0 : ℝ) → q (f n) x ≠ (0 : ℝ) → f (n + (1 : ℕ)) x = p (f n) x / q (f n) x) ∧ s = {x : ℝ | ∃ (l : ℝ), Tendsto (fun (n : ℕ) => f n x) atTop (𝓝 l)} ∧ ∀ x ∈ s, Tendsto (fun (n : ℕ) => f n x) atTop (𝓝 (g x))) := by sorry

\end{lstlisting}
\end{tcolorbox}

The answer type is:
\begin{tcolorbox}[colframe=codebg, colback=white, coltitle=blue!10!black, title=Lean, sharp corners=south, breakable]
\begin{lstlisting}
((((ℝ → ℝ) → (ℝ → ℝ)) × ((ℝ → ℝ) → (ℝ → ℝ))) × ((Set ℝ) × (ℝ → ℝ)))
\end{lstlisting}
\end{tcolorbox}
From this answer type, the dataset preprocessing mechanism infers a tuple answer type that contains a set. Since the source is a Putnam problem, the policy requires advanced arithmetic operators.
Therefore, the mechanism adds the allowed constants for advanced arithmetic, set-building, and tuple-building, resulting in the following admissible vocabulary:

\begin{tcolorbox}[colframe=codebg, colback=white, coltitle=blue!10!black, title=Lean, sharp corners=south, breakable]
\begin{lstlisting}
[And, Dvd.dvd, EmptyCollection.emptyCollection, Eq, Even, Exists, False, Fin.val, Finset.Icc, Finset.Ici, Finset.Ico, Finset.Iic, Finset.Iio, Finset.Ioc, Finset.Ioi, Finset.Ioo, Finset.card, Finset.product, Finset.range, GE.ge, GT.gt, HAdd.hAdd, HDiv.hDiv, HMod.hMod, HMul.hMul, HPow.hPow, HSMul.hSMul, HSub.hSub, Iff, Insert.insert, Int.ModEq, Int.cast, Inter.inter, LE.le, LT.lt, List.cons, List.nil, Membership.mem, Multiset.replicate, Nat.ModEq, Nat.Prime, Nat.cast, Nat.digits, Nat.factorial, Ne, Neg.neg, Not, Odd, OfNat.ofNat, Or, Prime, Prod.fst, Prod.mk, Prod.snd, Rat.cast, Real.exp, Real.log, Real.sqrt, Set.EqOn, Set.Icc, Set.Ici, Set.Ico, Set.Iic, Set.Iio, Set.Ioc, Set.Ioi, Set.Ioo, Set.Mem, Set.image, Set.insert, Set.singleton, Set.univ, Singleton.singleton, Subtype.mk, Subtype.val, True, Union.union, abs, ite, setOf]

\end{lstlisting}
\end{tcolorbox}

During the validation phase, the checker confirms that the above admissible vocabulary covers the constants used by the ground-truth answer, as shown below:

\begin{tcolorbox}[colframe=codebg, colback=white, coltitle=blue!10!black, title=Lean, sharp corners=south, breakable]
\begin{lstlisting}
[GE.ge, HAdd.hAdd, OfNat.ofNat, Prod.mk, Real.sqrt, setOf]
\end{lstlisting}
\end{tcolorbox}

\section{Prompt}
\label{app:prompts}
\subsection{Prompt for Autoformalization}

\begin{tcolorbox}[colframe=codebg, colback=white, coltitle=blue!10!black, title=Prompt, sharp corners=south, breakable]
\lstset{style=simple}
\begin{lstlisting}
You are given a math problem that requires answer-construction.
Your task is to formalize the problem in Lean 4 (v4.24.0).
1. Start with imports and namespaces such as `import Mathlib` and `open Real`.
2. Define the ground-truth answer as an abbrev for parsing:
   `abbrev <problem_name>_answer : β := <expression>`
   or
   `abbrev <problem_name>_answer : α → β := fun x => <expression>`
   (Don't put `(n : ℕ)` before `: α → β`; encode inputs inside `fun x => ...`.)
3. Formalize the theorem skeleton (leave proof as sorry).
   `theorem <problem_name> (x : α) (hx : P x) (y : β) :
      Q(x, y) ↔ y = <problem_name>_answer x := by sorry`
Other suggestions:
- Use `IsLeast` / `IsGreatest` for optimal answers.
- Use `Set β` and membership if there are multiple answers.
Your task: {Raw Problem}
\end{lstlisting}
\end{tcolorbox}

\subsection{Prompt for Enumerate and Conjecture Stage}
\begin{tcolorbox}[colframe=codebg, colback=white, coltitle=blue!10!black, title=Prompt, sharp corners=south, breakable]
\lstset{style=simple}
\begin{lstlisting}
I have a difficult high-school competition-level math problem in Lean, which asks for an answer with informal mathematical reasoning.

Your task is to attempt the problem and state the final answer as a Lean expression, with the help of Python function calls.

You are encouraged to use the tool code_exec to write code to enumerate the answer and obtain output, reason about the problem, and, if your reasoning and the code output disagree, keep doing the code/reasoning refinements until you
are confident with your reasoning steps and come up with a proposed answer that agrees with the enumerations. Be sure to use a print command to obtain the answer in Python.

Nevertheless, do not solely rely on the program enumeration. You should also provide mathematical reasoning for the answer.

Finally, you should provide a Lean answer and enclose your answer within the delimiter <<< >>> so I can parse it. You should include only the proposed answer in the delimiter (such as <<<1>>>), and you should never include complete declarations in the delimiter (such as `abbrev <answer_name> : <type> := 1`). You should not put proofs or justifications inside <<< >>>. It should include only the final answer.
Example: {Example}
Your task: {Raw Problem}
Previously failed answers: {Failed Answers}
\end{lstlisting}

\end{tcolorbox}
\subsection{Prompt for Proof Stage}
\begin{tcolorbox}[colframe=codebg, colback=white, coltitle=blue!10!black, title=Prompt, sharp corners=south, breakable]
\lstset{style=simple}
\begin{lstlisting}
Complete the following Lean 4 code:
{theorem statement}

\end{lstlisting}
\end{tcolorbox}


\end{document}